\pdfoutput=1 %For ArXiv submission
\documentclass[journal]{IEEEtran}

\usepackage{cite}
\ifCLASSINFOpdf
  \usepackage[pdftex]{graphicx}
  \graphicspath{{../img/}}
\else
\fi

\usepackage{algorithmicx}
\usepackage{algpseudocode}
\usepackage{enumerate}
\usepackage{fixltx2e}
\usepackage{stfloats}
\usepackage{latexsym}
\usepackage{url}
\usepackage{booktabs}
\usepackage{siunitx}
\usepackage{multirow}
\usepackage{color}

\begin{document}
\title{Using Behavior Objects to Manage Complexity in Virtual Worlds}

\author{
Martin \v{C}ern\'{y},
Tom\'{a}\v{s} Plch,
Mat\v{e}j Marko,
Jakub Gemrot,
Petr Ondr\'{a}\v{c}ek,
Cyril Brom% <-this % stops a space

\thanks{M. \v{C}ern\'{y}, T. Plch, J. Gemrot and C. Brom are with the 
Faculty of Mathematics and Physics, Charles University in Prague, Czech Republic.
E-mail: \{cerny.m,tomas.plch,jakub.gemrot\}@gmail.com, brom@ksvi.mff.cuni.cz
}% <-this % stops a space
\thanks{T. Plch, M. Marko and P. Ondr\'{a}\v{c}ek are with Warhorse Studios, Prague, Czech Republic.}% <-this % stops a space
\thanks{Manuscript received XX; revised XX.}
}


%\markboth{Journal of \LaTeX\ Class Files,~Vol.~11, No.~4, December~2012}%
%{Shell \MakeLowercase{\textit{et al.}}: Bare Demo of IEEEtran.cls for Journals}

% You can use \ifCLASs-objectPTIONpeerreview for conditional compilation here if


\maketitle

\begin{abstract}
The quality of high-level AI of non-player characters (NPCs) in commercial open-world games (OWGs) 
has been increasing during the past years. 
However, due to constraints specific to the game industry, 
this increase has been slow and it has been driven by larger budgets rather than adoption of new complex AI techniques. 
Most of the contemporary AI is still expressed as hard-coded scripts. 
The complexity and manageability of the script codebase 
is one of the key limiting factors for further AI improvements. 
In this paper we address this issue. 
We present \emph{behavior objects} --- a general approach to development of NPC behaviors 
for large OWGs.
Behavior objects are inspired by object-oriented programming 
and extend the concept of smart objects.
Our approach promotes encapsulation of data and code for multiple related behaviors in one place, 
hiding internal details
and embedding intelligence in the environment. 
Behavior objects are a natural abstraction of five different techniques 
that we have implemented to manage AI complexity in an upcoming AAA OWG. 
We report the details of the implementations in the context of behavior trees 
and the lessons learned during development. 
Our work should serve as inspiration for AI architecture designers from both the academia and the industry.

\end{abstract}


\IEEEpeerreviewmaketitle

\section{Introduction}
\IEEEPARstart{A}{} 
 growing number of computer games advertise to feature a 
``large open world''. 
No strict definition exists as whether a particular world 
can be considered ``large'' and ``open'' but one of
the key properties is definitely freedom: 
In an ideal case, the player is constrained only by the physical laws
of the virtual world 
--- they may interact at any time with all the objects and characters in their surrounding
and will always get a meaningful feedback from the environment.

Contemporary game worlds that are considered large feature
a landscape of tens to hundreds of square kilometers. 
% Recent and popular open world games such as Skyrim \cite{skyrim} 
% or Read Dead Redemption \cite{reddeadredemption} are actually in the
% lower part of the spectrum because of gameplay considerations --- 
% travelling through the world should not take too much time.
Such worlds are then populated with dozens of non-player characters (NPCs) that are
part of the story of the game and possibly hundreds of NPCs as
background cast.

% An important measure of quality of a virtual world is its believability.
% While the visual fidelity of contemporary games is spectacular, 
% this is not the case for NPC behaviors.

Development of the action-selection mechanism for NPCs in open-world games (OWGs) provides
both theoretical and practical challenges for applied AI.
Since the user has a large degree of freedom, the behaviors must not only
look meaningful to a spectator, but they must also maintain \emph{interactive believability},
i.e., NPCs should respond to the player's actions in a believable way.
While combat behavior in contemporary games is usually well designed and 
interactively believable to a large degree,
even recent and successful OWGs such as
Red Dead Redemption \cite{reddeadredemption} or Skyrim \cite{skyrim} have resorted to severely limited non-combat NPC behaviors.
Improved believability of non-combat behaviors would let the player feel like the world does not
revolve around them and increase immersion.
 
There are multiple reasons why games often implement only very basic non-combat behaviors,
but one of the most prominent is complexity:
the increased amount of behaviors an NPC is required to manifest in various contexts
results in a large and hard-to-maintain script codebase.
At the same time a shift from scripted behaviors to more intelligent, 
autonomous NPC design is currently not possible, 
as there are many NPCs to simulate and computing time is scarce. 
Similarly to the evolution of classical programming languages,
new scripting techniques are necessary to break the complex behaviors and their interactions down
into smaller manageable pieces \cite{complexityInHalo2}.

The state of the art in OWG AI scripting are --- to our knowledge --- 
variants of \emph{behavior trees} (BT) \cite{champandard2007behaviortrees}.
The common denominator of all BT approaches is that behavior is a tree structure 
which is traversed for every update of the NPC to determine an action (leaf) that should be executed.
The internal nodes then direct how the tree is traversed based on state of the nodes and input from the environment.
Conceptually the nodes close to the root correspond to high-level decisions while nodes close to the leaves
correspond to low-level decisions.

Other notable scripting techniques in use are finite state machines \cite{fu2004-fsm} 
and direct use of an interpreted procedural language like Lua \cite{luabook}
or combinations of the aforementioned techniques (e.g., BTs with Lua scripts as leaves).
In all cases we are aware of, their limitations are similar to those of BTs.

Many BT implementations provide techniques similar
to structured programming.
They allow for subtrees to be reused across multiple NPCs --- a form of ``behavior subroutines''.

During our work on AI system for an AAA OWG,
we implemented an augmented variant of BTs,
but more complex non-combat behaviors were still very challenging to develop.
Therefore we designed and implemented several new scripting techniques to reduce the complexity of the BT codebase. 
The common denominator of all the techniques 
was the encapsulation of related behaviors and associated data
into a unified structure (e.g., behaviors for innkeeper and guests in a pub along with reference to chairs in the pub).
We have noted that this is conceptually similar 
to the object-oriented programming paradigm (OOP). 
In this paper we present the concept of \emph{behavior objects}
as a natural abstraction of the scripting techniques we implemented.
We describe the individual techniques as instances of behavior objects
%and explicitly link behavior objects to the OOP paradigm.
and note specifics of behavior development 
for which direct application of OOP concepts would be problematic 
and describe how we accounted for those specifics.

Although our work is based on BTs,
behavior objects are more general and can be directly transferred to many other NPC AI approaches 
that are common in the industry.

This paper is an extension of our papers on smart areas \cite{cerny2014smartareas} and scheduling of small cooperative behaviors \cite{situationsystem}.
The novel contribution in this paper is 
a) the description of three previously unpublished techniques that use similar encapsulation philosophy as smart areas,
b) synthesis of general insights from the individual use cases resulting in the concept of behavior objects
and  c) thorough discussion of qualitative feedback from the game designers and scripters.

The paper is organized as follows: 
first, we introduce the problem of behavior development in greater detail (Section \ref{sec:problem})
and discuss related work (Section \ref{sec:related}).
Next, we analyze the requirements for the system (Section \ref{sec:analysis}), introduce behavior objects as our proposed solution (Section \ref{sec:solution})
and describe our particular implementation of behavior objects (Section \ref{sec:impl}).
Last we present qualitative evaluation of our implementation in practice (Section \ref{sec:evaluation})  
and conclude the paper (Section \ref{sec:discussion}).

\section{The Problem}
\label{sec:problem}
This section describes
the constraints for AI in OWGs in general 
and discusses the problem we aimed to solve.
It is based on the experience gained
during our collaboration
with a game studio that is developing a high-budget  
OWG and on casual discussions with game developers at various conferences.
While definitely subjective, we believe that most of the following claims
generalize to the industry as a whole.
% First we discuss individual aspects of the AI that need to be dealt with in an OWG
% and summarize general constraints on game AI. 
% Then we detail the problem statement and provide basic analysis.
% Since our game is set in medieval Europe, 
% the examples we provide are taken from this environment.

\subsection{Believable Behaviors}
OWGs strive to maintain suspension of disbelief on the player's side.
%The player should experience feelings of freedom and agency, 
%they should feel that they entered a world where their actions are meaningful.
A well-executed game lets the player forget that the game world is imaginary
and lets them immerse in the experience. 
%If the player sees NPCs as ``real'' characters, 
%it is much easier for them to create emotional bonds with them,
%letting the player construct their own story in the world.
While the player is immersed in the game, they are more likely to act
in-character, to experience the world as its inhabitant rather than an external observer. 
NPC behaviors can make a huge difference on this part:
an NPC that repeats a phrase the player has heard many times from others 
(as in the Skyrim's \cite{skyrim} famous ``Arrow in the knee'' dialogue\footnote{\url{http://knowyourmeme.com/memes/i-took-an-arrow-in-the-knee}}), 
an NPC that stays at the same place all day or walks through a door,
or simply an inappropriate animation may break the immersion quickly.
Without immersion, NPCs start to appear more like puppets
whose sole purpose is to issue quests or provide items the player can steal;
interactions with NPCs lose meaning and become a mechanical obstacle to overcome.
Improving believability of NPC behaviors helps maintain the suspension of disbelief and thus enrich the player's experience.

\subsection{Game AI Components} 

In a typical OWG, such as Grand Theft Auto \cite{gtaV} or Skyrim, the NPC AI may be divided into two main components.
As fighting enemies is still a major part of most OWGs, \emph{combat AI} is often the largest AI subsystem. 
%   It may be further divided into \emph{enemy AI} that guides NPCs opposing the player and \emph{ally AI}
%   that controls NPCs trying to help the player in a fight.
\emph{Non-combat AI} governs the rest of the NPC behavior. It may be further divided into direct interactions with the player
  (e.g., dialogues, barter, \ldots) and \emph{ambient AI} which covers the daily life of the NPCs and other actions they perform on their own.
  Ambient AI may greatly increase the appeal of the world and make it appear more alive to the player, yet is currently arguably the least developed.
  Our main focus is on improving ambient AI.    
% In a particular game the individual components may be further subdivided and other components may be
% added to suit the needs of the game (e.g., a component that coordinates groups of NPCs).

\subsection{Industrial Constraints}
\label{sec:industrial_constraints}
There are three main challenges for OWG AI in general. 
First, in AAA computer games CPU time is a very scarce resource
as almost all of the CPU time is dedicated to graphics and physics.
In our project we got up to 5 ms per frame (100 - 150 ms per second) on a single core
for all NPCs together, including pathfinding and collision avoidance.
Even if the system would simulate only the NPCs in direct visibility to the player (might be a few dozen)
it is left with several milliseconds per second and character 
for the actual action selection.
This disqualifies computationally expensive techniques such as AI planning from use in ambient AI,
despite their usefulness in small-scale scenarios such as combat \cite{champandard2013planningoverview}.

Second, the game industry needs to exercise tight control over NPC behaviors.
Unpredictable or hard-to-understand behavior 
is usually undesirable as it may break
gameplay or interfere with the flow of the game's story. %as envisioned by the designer.

Third, the industry is also concerned with effective (in terms of time and money) development of the AI.
Designing a large number of complex behaviors thus introduces problems known from software engineering:
code readability and reusability and management of the development cycle.
As it is cost-prohibitive to hire 
expert programmers for behavior design, the technology used should also be relatively simple.
% It is also important to consider that it is cost-prohibitive to hire 
% expert programmers for behavior design (those usually work on engine core) 
% so the designers usually have little programming skills.
% Therefore, using a complicated technology (e.g., an agent-oriented programming language)
% is not an option, unless high-level tools can be provided \cite{gemrot2014methodology}.

Due to those constraints,
scripted approaches are still state of the art for non-combat behaviors in large OWG scenarios
and will likely remain so in the near future.
Variants of the BT formalism are to our knowledge by far the most frequent.
The supplemental material for this paper provides concrete examples of shortcomings of state-of-the-art ambient AI.

\subsection{Problem Statement}

The problem is that to increase the perceived complexity of scripted behaviors beyond the state of the art,
more powerful approaches to managing behavior code complexity are needed.
As in classical programming, separation of concerns and hierarchical decomposition are vital for good code structure.
In particular, behaviors must be allowed to execute sub-behaviors (e.g. reuse code for sitting on a chair within a behavior for drinking in a pub).
The nature of our project introduced additional design objectives the new technique should fulfill:

\begin{enumerate}[{\textbf{O}1:{}}] 
  \item \label{problem:gameplaycritical} Strong guarantees must be made that gameplay-critical behaviors (quests, combat, \ldots) 
will not be disrupted.
	\item \label{problem:consistency}  The behaviors must be interruptible and maintain consistency even on prolonged execution (i.e., dozens of hours). 
	%The game is expected to be played for several dozen hours while all the NPCs are continually simulated, without any reset.
  \item \label{problem:ambient} Primary use-case is the ambient AI.
  \item \label{problem:decoupling} The behavior code has to be decoupled from the data in the game world. 
  In particular, using an already defined behavior in a new context (e.g., adding a new pub to the game world) should be possible
without changing any of the code of NPCs that use the behavior and without modification to the pub logic.
  \item \label{problem:different} Some NPCs should be allowed to behave differently in the same context:
e.g., in a pub, rich people behave (and are treated) differently than poor people. 
  \item \label{problem:synchronization} Coordination of joint behaviors among agents (a pub brawl, a game of cards, \ldots) must be
  supported. 
\end{enumerate}

To an extent, all of these additional objectives can be achieved with state-of-the-art scripting,
but at the cost of increased code complexity. 
Our aim is to meet our objectives while reducing code complexity at the same time.  

\section{Related work}
\label{sec:related}
\noindent 

In both academia and industry, a prominent approach to managing behavioral complexity
is embedding intelligence in the environment. 
The most common example are 
\emph{smart objects} as introduced by Kallmann \cite{kallmann2002smartobjects_paper}, 
although other approaches also exist (see below).
Smart objects are, to our knowledge, well-established in the game industry, although in a very simplified form. 
A smart object as used by the industry is typically a graphical entity in the game world 
that is accompanied by a character animation (or several animations) 
that should be used when a character desires
to use the object.
The smart object is also responsible for positioning the character at 
the exact spot where the animation should be played.

A typical example of a smart object is a lever on the wall. 
An NPC that wants to change the state of the lever simply fires
a ``use smart object'' action and the smart object takes care of the necessary
details. 
This way, many different levers and switches may be present in the environment,
but the AI only needs one action to use them all properly.

Another frequent use are so-called \emph{navigation smart objects},
which are smart objects intended solely for navigating around the environment.
A navigation smart object connects a graphical entity in the game world with an entry and an exit point.
When an NPC wants to move from the entry point to the exit point, it plays the animation associated 
with the smart object.
Navigation smart objects typically connect disjoint areas that could not be traversed by regular navigation 
--- a typical example is jumping over a barrier. 
% Another approach to handle barriers and doors is to embed additional navigation information inside edges of the navigation graph \cite{dataInNavigationGraph-AiWisdom}.
% While this is conceptually similar, it is less flexible as the navigation graph needs to be manually kept consistent with the environment.

Kallmann originally proposed smart objects as more complex entities that can provide multiple  
interacting parts, each with its own location, mechanics, instructions for NPC positioning
and optionally also with code the NPC should run to use the given part. 
Kallmann's smart objects could also run code on their own to alter their internal state.
Kallmann's idea is close to our goal, 
although several important features are missing.
Most notably it does not consider interrupts to the behaviors,
and it does not support behavior nesting.

A version of smart objects close to the Kallmann's version has been implemented in The Sims series \cite{sims4gdc}. 
The Sims form a very different application than OWGs, because the user is not embodied in the environment and interactions with NPCs and objects are triggered indirectly.
The NPCs autonomous decision making consists of selecting an appropriate smart object (NPCs are also smart objects) to satisfy its current needs.
The basic unit of behavior in The Sims 4 is called ``interaction'' (e.g., sit down). 
Interaction consists of animations and changes to state of the NPCs and/or the world (e.g., NPC is now in ``seated'' pose, chair is occupied).
These interactions are then connected to objects in the game (e.g., the same sit down interaction is connected with a chair and a bench).
The interaction may further decompose into atomic ``blocks''. 
Those blocks are not interruptible and are always run to completion, but blocks of multiple interactions may be interleaved (e.g., sip a drink --- look at TV, cheer --- finish the drink)
and nested (e.g., cuddling while sitting on a sofa).

% Our work is the first that we are aware of that implements comparably powerful smart objects in a game where the user
% may interact directly with the NPCs. 

% Conceptually, smart objects are inspired by the psychological notion of \emph{affordances} \cite{gibson1986ecological}.
% The idea is that animals (and humans) do not perceive the environment as physical objects but rather
% as a set of possibilities the environment affords, e.g., a door is something that may be opened, lever is something to be pulled, barrier is something to jump over, etc.

Although our experience indicates that smart objects are used in
many first/third person games, there are --- as is often the case with game industry --- relatively few official sources and very little detail was revealed.
Among those, smart objects are mentioned in the context of FarCry 4\cite{farcry4},
Castlevania: Lord of Shadows\cite{castlevania} or F.E.A.R. 2 \cite{fear2}.
BioShock:Infinite also has ``opportunities'' placed in the environment for the sidekick character Elizabeth to interact with\footnote{\url{http://www.youtube.com/watch?v=2viudg2jsE8}}.
From the little information available, all these implementations do not seem to go much beyond levers and other simple objects. 

Notably the S.T.A.L.K.E.R. series extended smart objects to ``smart terrains'' that provide more long-term behaviors to all NPCs in a specific area \cite{champandard2008stalkerai}.

Another interesting approach is presented in Hitman: Absolution.
Here, the AI uses objects called ``situations'' to coordinate multiple NPCs \cite{hitmanAbsolution}.
% Whenever an NPC deals with an event that requires coordination with others (e.g., the player is trespassing and should be stopped), 
% it subscribes to a corresponding \emph{situation object}.  
The situation object assigns roles to the subscribed NPCs and alters their knowledge based on events in the game world 
%(e.g., tells the NPC that it is in a trespassing situation, who is the leader of the situation and how aggressively should the NPC react). 
(e.g., how aggressively should the NPC react). 
The NPCs then take that knowledge into account in their own decision making. 
The drawback of this approach is that every NPC needs to include specific
code for every situation it may participate in. 
Furthermore, the code for the situation is scattered among multiple NPCs.
%and all variants of the situation code have to be considered if an NPC 
% needs to coordinate its behavior with actions of the other participants.    
% Our approach has the advantage of decoupling situations from the rest of the AI and 
% it allows for tighter coordination among NPCs at the cost of fixing the number of NPCs for a situation.

With regards to the major available game engines, 
CryEngine Free SDK seems to have the best support for embedding intelligence in the environment\cite{crySmartObjects}.
In CryEngine Free SDK a ``smart object rule'' can be assigned to any entity in the game. 
The rule consists of a condition and an AI script to be executed, when the condition is met.
This approach allows for simple creation of a wide variety of active non-character entities (landmines, machines, \ldots),
but the approach is more problematic when providing behaviors for NPCs as the script within the rule executes in parallel with
the NPC's logic. 
In all but the simplest situations the script within the rule would have to manually synchronize/communicate
with the NPC's logic to prevent the rule from threatening the consistency of the NPC's state 
or from interrupting a gameplay-critical behavior,
introducing unnecessary coupling of the respective code. 
There is also no support for communication when multiple NPCs use the same object.
In general the ``smart object rules'' of CryEngine Free SDK are great tools for what CryEngine was intended for --- quick action in first-person shooters ---
but they are not very suitable for ambient AI in complex persistent worlds.  

Unity3D and Unreal Engine have no built-in support for embedding intelligence into the environment, 
although there are AI middleware solutions that provide some support.
One example is Autodesk Gameware Navigation\footnote{\url{http://gameware.autodesk.com/navigation}}
that includes support for navigation smart objects (as described above).

In academia, the concept of smart objects has been extended by crowd simulation
research to whole areas. 
In  \cite{tecchia2001abs} the environment is overlaid with
a grid, where each cell may dictate a movement behavior for the agents in it.
In \cite{sung2004scalablebehaviors} ``situation based behavior selection''
is presented. 
The system detects situations in the environment and instructs
the agents participating in the situation what should they do.
While situation based behavior selection is very general, 
the situations tested in the paper are mostly triggered by entering a location.

Stocker et al. \cite{stocker2010smartevents} introduced ``Smart Events'' --- 
specific objects providing NPCs with ready-made responses to external events.
An important problem is that Smart Events provide the same behaviors for all types of characters 
and do not provide means for coordination among the characters. 

The simulation of Shao and Terzopoulos \cite{shao2007autonomouspedestrians} 
features autonomous pedestrians in a virtual railway station. 
Several social behaviors (e.g., buying tickets, spectating an art show) with
coordination (e.g., queue at the ticket booth) mediated by specialized environment objects
are introduced.
However, every character must be explicitly
prepared for all the social behaviors it may perform, limiting scalability. 

Further, the crowd simulation approach cannot be directly translated to 
OWGs
% There are two reasons for this: 
%first, crowd simulations generally have low fidelity of individual behaviors
%and second, 
as crowd simulation is intended to be believable from a larger perspective,
but does not necessarily retain believability when individual characters are tracked. 
% To quote \cite{sung2004scalablebehaviors}:
% ``When we look at a crowd, we care only
% about what is happening, not who is doing it'' In their work this let them adopt 
% a simple probabilistic behavior selection model, which is unsuitable for computer games
% as it allows a character to e.g., go to work twice without a break.

Brom et. al. \cite{brom2006ive}  take the idea of smart objects further
with ``smart materializations''. 
In their work the world is inhabited by agents using the belief-desire-intention (BDI)
architecture.
The only way to act on intentions is to choose a smart materialization
which is a behavior fragment embedded in the environment. 
The smart materialization may in turn introduce subintentions, which are again
resolved in the same manner.
For example the character may adopt a ``have fun'' intention.
A pub in the environment would provide a materialization that realizes the 
``have fun'' intention by instructing the agent to go to the pub and
adopt subintentions ``buy a beer'' and ``drink a beer''.
A simple scheme to choose the best materialization among those that achieve the same intention
is implemented. This work has provided substantial inspiration for us.
 
While smart materializations have many of the desired properties, they lack the possibility to 
create behaviors or their parts without any materialization. 
Also BDI architecture is seldom used as a game AI architecture,
probably because it is relatively complex and not well known to the developers.

Orthogonally to embedding intelligence in the environment (or outside of NPCs in general),
Bryson \cite{bryson2001thesis} advocated using object-oriented programming principles in behavior design.
In her view, every capability of an agent should be represented as an object. 
Bryson's approach however has no explicit support for agent coordination neither does she outline
the use of objects of a finer granularity.

A different approach to modularizing behaviors is provided by ScriptEase \cite{scriptease}.
ScriptEase lets users create behaviors by using \emph{generative design patterns} (GDPs)
which are essentially parameterized code generators. 
This approach allows for great flexibility as the scripter can make low-level modifications to the generated
code but does not account for hierarchical decomposition of complex behaviors.

\begin{figure*}
\centering
\includegraphics[width=0.9\textwidth]{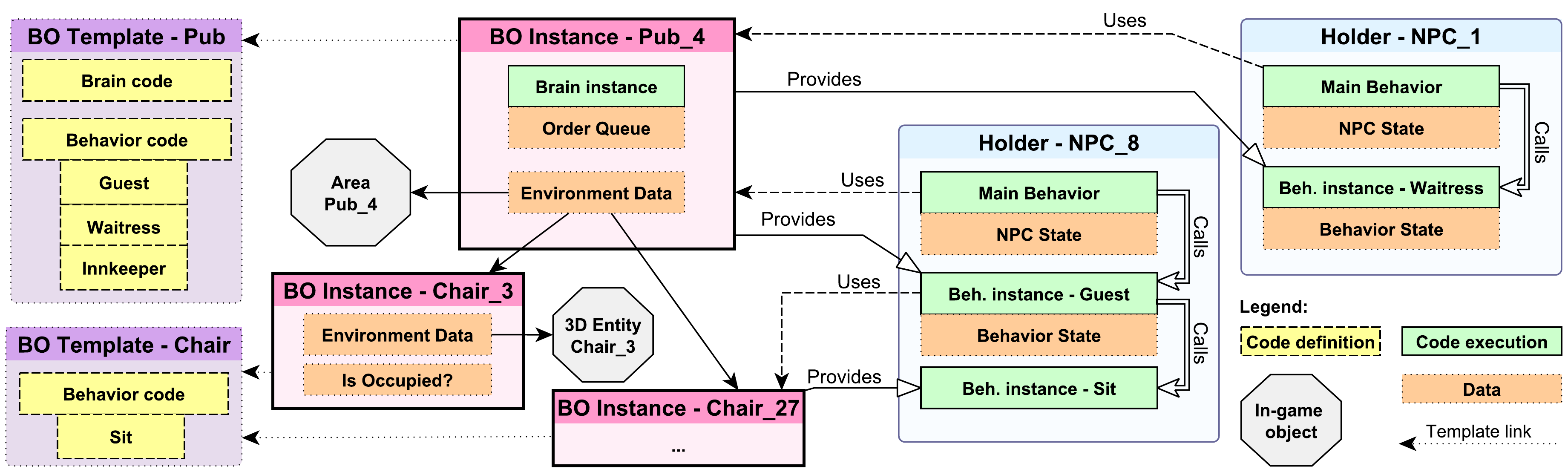}

\caption{An example usage of behavior objects: a pub with multiple chairs. The code for individual behaviors and central decision logic is provided in BO templates (purple, dotted), 
that are shared by multiple instances (pink, solid). The instances execute code and encapsulate state and environment data --- links to in-game entities and other BOs. 
NPC\_1 represents a waitress --- it uses ``Waitress'' behavior provided by the pub and manages data used by the behavior. NPC\_8 is a guest in the pub --- it uses the ``Guest'' behavior 
provided by the pub,
which further uses ``Sit'' behavior provided by a chair.}
\label{fig:bo_structure} 
\end{figure*}

\section{Analysis}
\label{sec:analysis}

The greatest complication in addressing the problems outlined in Section~\ref{sec:problem} is the need to maintain interactive believability to a reasonable degree. 
For example, if the user performs a violent act on the street,
the NPCs nearby should not keep walking to a pub, but should
rather run away scared. 
This in turn should result in few people arriving to the pub
which should affect the way innkeeper behaves.

While the actual execution of the behavior is relatively simple,
the possibility of interruption at any time 
and huge amount of edge cases to be handled (e.g., no innkeeper present at the pub)
makes the code grow substantially in size and complexity.
With a naive approach it would be very difficult 
to maintain, test and debug the resulting codebase.
% To our knowledge such level of behavioral fidelity 
% has never been achieved in high-profile commercial computer games.

% In context of ambient AI, the NPC does not have to express a story-making behavior. 
% Rather, it should play a behavior that fits an NPC's situation, 
% a combination of NPC's background, current environment and concrete situation as implied 
% by other story-making elements (e.g., the story, a quest, the player).
% It is the NPC that should be responsible for managing this behavioral context (e.g., when it works, what occupation it has),
% but the NPC should not be responsible 
% Therefore, designers need to have a way, how to break down the whole ambient character behaviors 
% into pieces and a way to specify how these pieces should be sewed back into the complete behavior during runtime.

% Therefore, designers need to be able to break down the whole ambient character behaviors 
% into pieces and to specify how these pieces should be sewed back into the complete behavior during runtime.

The key to managing complexity is modularization. 
% In a well modularized code, the behavior is split into small units
% which may be referenced as black boxes, without the need to understand their content (an ideal to pursue). 
A key prerequisite for proper modularization are behavior execution semantics that ensure that upon interruption,
the NPC remains in a well-defined state and that the interrupting behavior does not need to be aware of 
the exact state of the interrupted behavior. %(e.g., in the middle of complex animation sequence, e.g., drinking the beer).
Further, to make implementation of cooperative behaviors easier, it is useful 
to have explicitly shared data structures and/or communication channels so that 
the communication channels do not have to be negotiated at runtime. 

Some form of modularization is already present in state-of-the-art game AI systems.
It consists mostly of reusing behavior fragments across multiple behaviors. 
Using classical programming terminology, this is conceptually equivalent to calling subroutines in structured programming.
This however seems not satisfactory for our case as it would be difficult to properly connect behavior code (e.g., a behavior subroutine for a guest in a pub)
with necessary data (e.g., positions of chairs in the pub) and references to other NPCs (e.g., the innkeeper) without tightly coupling the code to a particular data
or creating ad-hoc methods for data acquisition.

We have used two main modularization approaches:
a) embedding intelligence into the environment
and b) taking inspiration from object-oriented programming (OOP).

Embedding intelligence into the environment means
that instead of the NPC knowing how to behave in a specific context (behaving at a location, using an object, \ldots)
the context itself provides behaviors for the NPCs.
The greatest advantage of embedded intelligence is that it allows adding new contexts to the game world
without modifying the NPC code (as in objective O\ref{problem:decoupling}).
 
OOP was our inspiration in multiple ways. 
It naturally connects code and data and provides encapsulation of code into larger structures.
OOP promotes hiding of internal information from 
unintended manipulation from the outside.
OOP is also connected to well-tested methods to handle coordination among multiple threads, 
which relates closely to coordination among multiple NPCs.
Last but not least, OOP is accompanied with methodology of decomposing a large problem into self-contained
objects that then delegate specialized work to each other and can be tested independently. 
  
On the other hand, as most of the behaviors we considered have only very small fractions in common,
implementing inheritance to ease code sharing among behavior objects is not necessary and would only complicate things further.
A simple form of polymorphism is however desired;
in particular it is useful to
allow the NPC to use a given behavior from different objects that provide the behavior.
For example, the NPC may want to perform a ``have fun'' behavior  
without caring whether the object being requested for the behavior represents a pub or a playground. 
The simplest way to achieve this is to let any behavior be requested from any object as long as the object has the behavior defined.
In OOP context this practice is called ``duck typing'' and is frequently used in scripting languages.
% Polymorphism without inheritance is very easily achieved by using ``duck typing'' for the behavior objects:
% any behavior can be requested from any object as long as the object has the behavior defined.
% Duck typing aligns nicely with game AI environments 
% as behaviors are usually represented in a dynamically typed scripting language.

\section{Proposed Solution --- Behavior Objects}
\label{sec:solution}

\emph{Behavior objects} (BOs) are the behavioral parallel to object-oriented programming (OOP).
Objects in OOP consist of code (methods) and data (fields).
The code is defined once for a class of objects, while data are specific to object instances.
When a method is invoked on an object, it manipulates the object's data to provide a desired result.

BOs consist of code (behaviors), data and central decision logic which we call the \emph{brain}.
Code and brain is defined in a BO \emph{template}, the data is specific to a BO instance
(this addresses the objective O\ref{problem:decoupling}).
When an NPC uses a behavior provided by a BO, it executes the behavior in its own context
and lets it access the NPC's internal state.
The NPC becomes a \emph{holder} of the behavior.
The behavior however still has access to the BO instance data which provide further context for execution
and provide an implicit communication channel to other NPCs using behavior from the same BO instance (addressing objective O\ref{problem:synchronization}).
The brain (if present) manages the individual behaviors
and may actively influence their execution, either by manipulating the BO instance data or by explicit communication with NPCs
holding BO's behaviors.
% This puts BOs closer to intelligent agents, 
% but BOs are conceptually very different from agents --- they are not independent and their purpose is to primarily
% provide behaviors to the NPCs.
The BO instance data come in two very different flavors: \emph{environment data}, which are links to entities in the game world, 
and \emph{state}, which is internal to the object.

A simple example of a BO is a chair with a ``sit'' behavior --- 
here the environment data consists only of the chair; the object state is a flag indicating whether the chair is in use,
 and the behavior consists of three animations (sitting down; idle while sitting; standing up).
 The chair has no brain.
A complex example is a BO that manages a pub. 
It contains behaviors for guests, the innkeeper and the waitress.
The environment data consist of links to chair BO instances (as above) inside the pub and the area the pub covers; 
the state is a list of orders for drinks.
See Figure~\ref{fig:bo_structure} for a diagram of the situation.

The brain of the pub BO handles requests for seats and drinks
and sends messages to guests to inform them where to sit and to the innkeeper and the waitress to instruct them to
prepare and deliver drinks respectively.
This makes the pub a central point of communication,
which allows multiple waitresses/innkeepers to be added to the pub without changing the code.
Since the code for guests, the innkeeper and waitress is all in the BO, 
the development of the communication protocols for seat and order management is greatly simplified
as all its uses are from within the BO. 
To use OOP terminology, the communications are private to the pub object. 
Most of the rules of thumb used in object-oriented analysis can be easily translated to the behavior case 
to help design a good decomposition of behaviors into BOs.

\subsection{Differences from OOP}

We found three notable issues specific to behavior development 
that preclude direct application of classical OOP approach.
These issues motivate the main differences between BOs and OOP.  
First, the state of an NPC
is implicitly shared by all behaviors the NPC may execute, which complicates encapsulation.
Second, behaviors have different execution model than programs.  
Third, execution of behaviors is highly parallel --- all NPCs and BO brains act like separate threads.
This section details these three issues and how we addressed them with BOs.

%kandidat zase tolik to nesouvisi s BO 
\subsubsection{Shared State}
\label{sec:shared_state} %Pozor, na tohle je odkaz!
The fact that the NPC's state (position, speed, active animation) is shared by multiple behaviors makes encapsulation of code more difficult.
As a general rule, a behavior should terminate only when all the changes to NPC's state have been fully
completed or rolled back and a behavior should always check the NPC's state anew when it resumes execution after an interrupt.
This is well supported by the underlying BT formalism.

The reality of the game engine has forced us to make an important exception to the above rule:
Since a behavior may require an extensive computation or data exchange with other NPCs/BOs
to determine the next action, the system cannot guarantee that a new behavior will issue an action on the same frame in which the old behavior has ended.
This would result in movement and animation artifacts where the NPC stays still for one or two frames during a behavior transition. 
To remedy this, a behavior should terminate before its last movement and/or animation action completes. 
Every behavior is then required to issue an animation and/or a movement action (or force the NPC to stop)
at the beginning of its execution. 
This way, the transitions are instantaneous and 
the animation subsystem can take both new and old animations into account when choosing an appropriate transition animation.
As almost all behaviors start with movement or animation anyway, 
this approach required very little modification to behavior code
and worked reasonably well in practice. 

\begin{figure}
\centering
\includegraphics[width=\columnwidth]{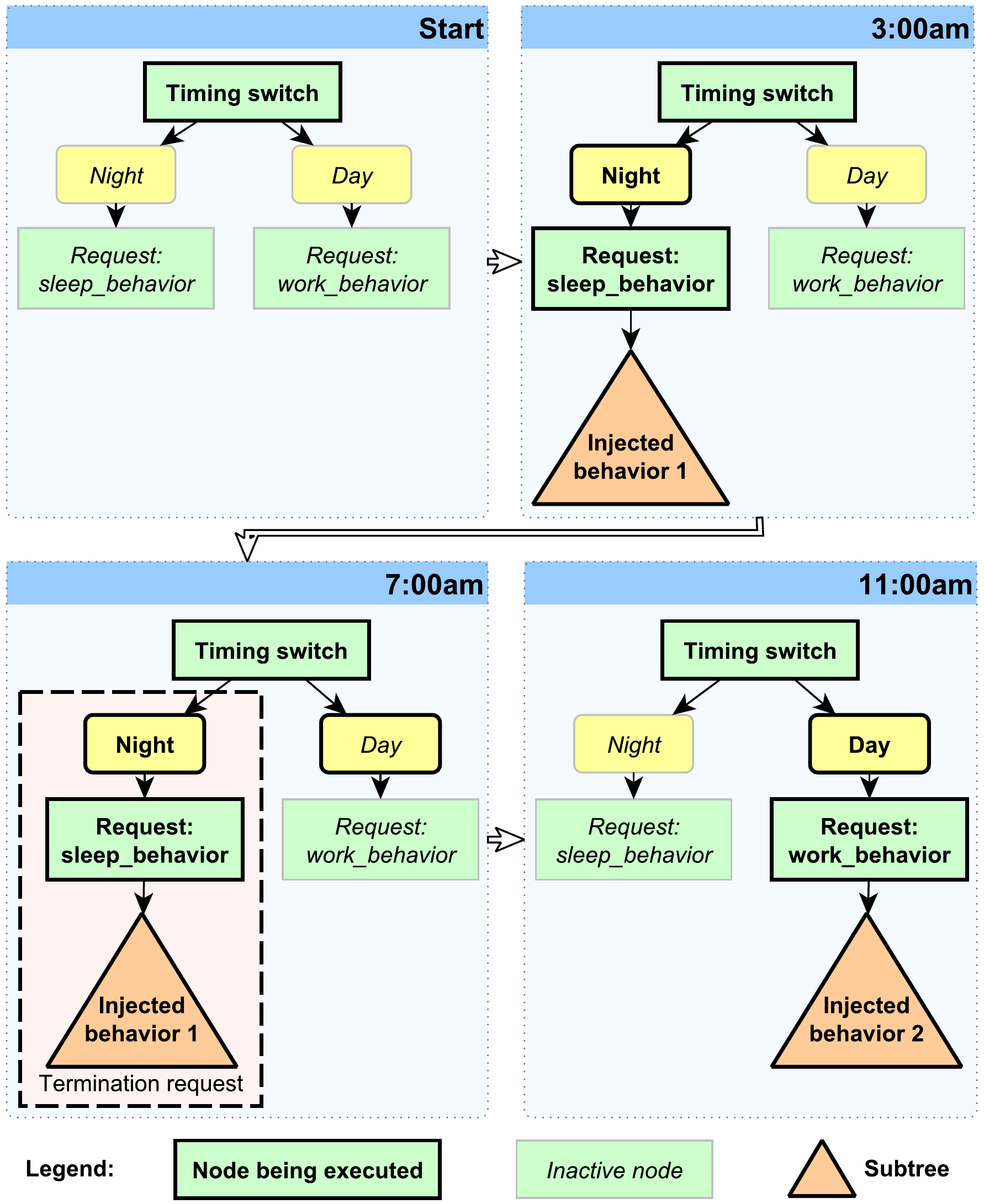}

\caption{Injecting behavior into an NPC's main behavior.
In this case, it is the NPC that actively requests a behavior of a given type to be injected. 
Since the higher-priority nodes are still being evaluated when performing the injected behavior, the high-level NPC decision making
may terminate the injected behavior when necessary. 
A similar logic handles the ambient AI of NPCs in our game. }
\label{fig:injection_termination}
\end{figure}

\subsubsection{Execution of Behaviors}
In OOP, executing a method results in a full change of context 
--- the methods lower on the stack do not influence the execution of the method.
Behaviors are different, because they are \emph{injected}: using a behavior of a BO results in
inserting a new subtree in the BT that drives the NPC 
and thus the parent behavior still influences execution.
In particular a node closer to the root may switch 
to a different child and terminate the injected behavior --- see Figure~\ref{fig:injection_termination} for an example.
% The behavior may also be injected at a well-defined place in the NPC decision making that  
% is not the part of the main BT. 
% For example, the NPC architecture may allow asynchronous execution of a small ``event'' tree whenever the NPC is damaged.
% The behavior then may be injected as this event tree.
Note that fully replacing the current behavior of the NPC is not a viable option 
as the BO behavior would then need to be able to react to high-priority external events (e.g., combat) on its own,
reducing modularity of the code.  

We have considered two flavors of behavior injection:
Either it is \emph{on-request} --- NPC behavior actively requests a behavior from a BO 
and is in direct control of the injection
(as in the example), or \emph{on-command} --- 
a behavior is imperatively injected into NPC behavior structure based on conditions external to the NPC (e.g. injecting code to handle a combat event).
Nevertheless, even the on-command injection still keeps the top-level NPC logic in-place and it is the top-level logic which
decides when and if at all the injected code is executed (this helps to maintain consistency --- objective O\ref{problem:consistency}).
For most of our use-cases, the on-request method is more appropriate: 
it is the NPC that decides to perform a work behavior 
and the actual request for the behavior should be made at the time of such decision.
However, we use on-command injection as well. 

Note that the injection principle can be applied to formalisms other than BTs.
For example, when finite-state machines (FSMs) \cite{fu2004-fsm} are used, 
a special ``use behavior object'' state may be expanded to a new FSM prior to transition to that state.
For a belief-desire-intention architecture, methods to act on certain intentions may be injected, etc.
However, BTs naturally provide a very clean support for decomposition and hierarchical 
structuring of behaviors which aligns nicely with the BO approach.

\subsubsection{Parallel Execution}
While OOP languages provide mechanisms to handle parallelism,
BOs differ in the scale of the problem --- effectively every 
NPC and BO instance acts as a separate thread, 
and thus parallelism has to be accounted for at the architecture level and not ad-hoc
at the code level. 

The main problem arising from the parallel nature of game AI is safe and consistent data access
and sharing required for coordinated behavior (O\ref{problem:synchronization}).
While it is safe to directly access immutable data belonging to a different thread, 
access to mutable data needs to be more careful to avoid race conditions.
There are multiple solutions to this problem in OOP,
but we consider it best to make each thread (NPC, BO brain) solely responsible for its mutable data.
Mutable data of other threads can be accessed only indirectly by sending messages to
a) request data from another thread, b) provide data to another thread or c) request a change of data belonging to another thread.
The receiver then handles those messages within its own updates.
Using the message system as the sole mechanism
for sharing mutable data 
% prevents a large amount of race conditions 
% and other issues that could arise from interleaved execution of BO behaviors and the brain.  
% In most cases, processing messages one by one results in a code that is robust to any possible interleaving with other behaviors. 
is in most cases sufficient to ensure that the behaviors are robust to any possible interleaving with other behaviors.
  
\begin{figure}
\centering
\includegraphics[width=\columnwidth]{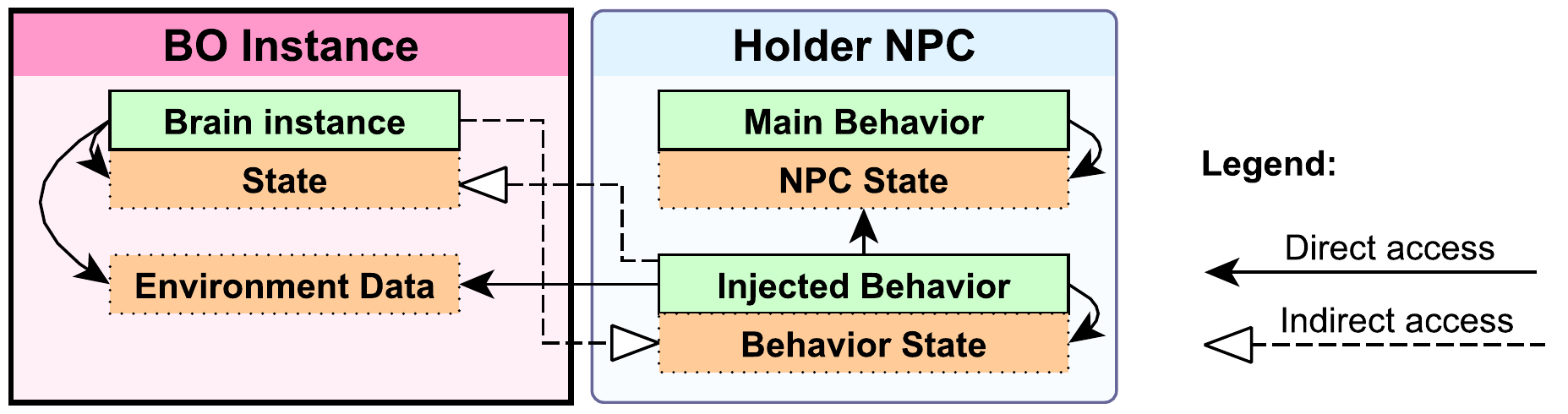}

\caption{Possible data access between a BO instance and a holder. In direct access the data may be explicitly referenced, while indirect access requires sending messages to read/write data.}
\label{fig:data_access}
\end{figure}

In most cases, BO's environment data is immutable at runtime and thus may be directly referenced from anywhere.
Internal state on the other hand is almost always mutable and thus cannot be referenced directly from other threads.
Since an injected behavior is executed within the main behavior of an NPC, it has full access to the NPC's internal state
but the state of the injected behavior cannot be directly referenced by the BO's brain and vice versa (see Figure~\ref{fig:data_access}).

Like OOP, the BO approach is not a silver bullet to solve all behavior design problems,
but it has the potential to mitigate complexity and enable scripters to create more lifelike behaviors within a given timeframe.

\section{Implementation}
\label{sec:impl}
\begin{figure}
\centering
\includegraphics[width=0.65\columnwidth]{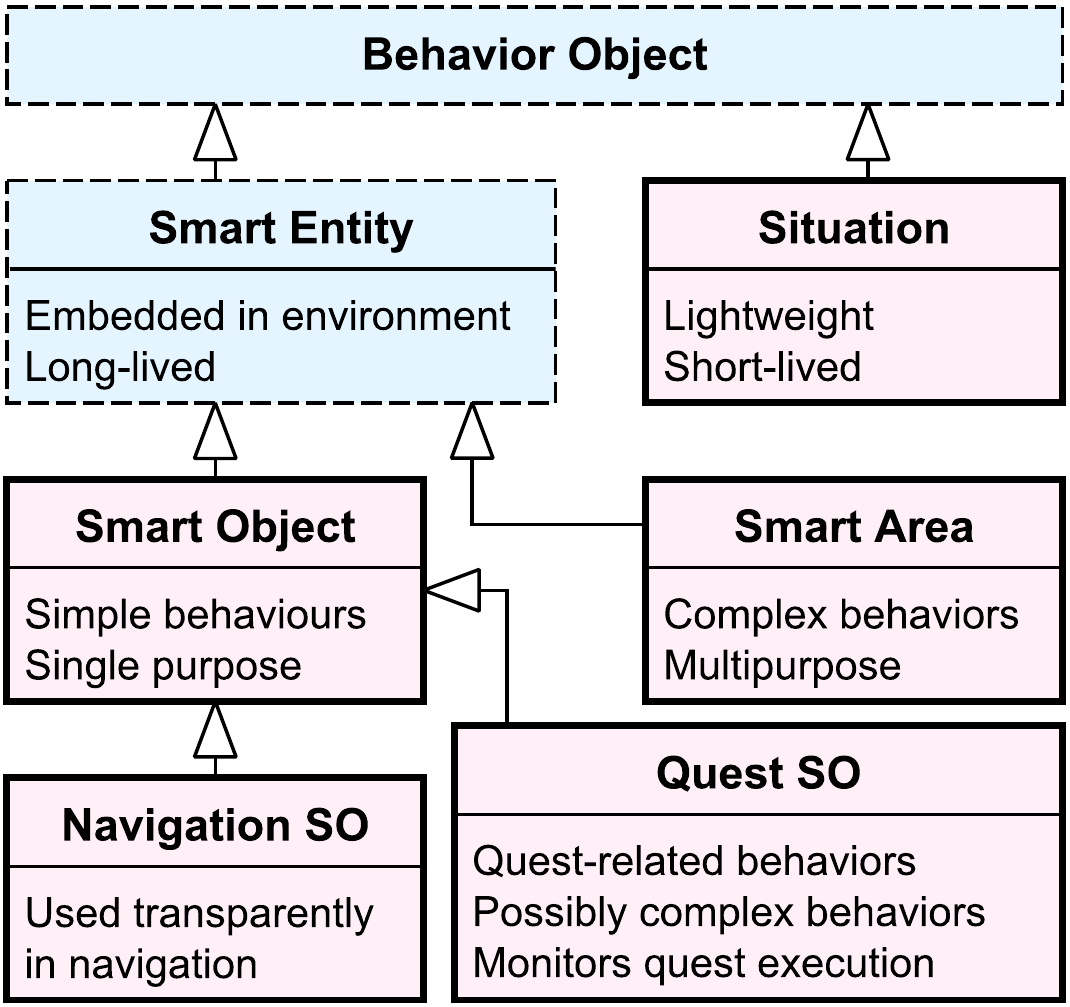}

\caption{A simple class diagram of various types of behavior objects and their basic properties. Abstract categories have dashed borders, 
while full borders correspond to types of BOs that are actually used in the game.}
\label{fig:objects_overview}
\end{figure}

\noindent In this section we describe the
implementation of BOs for our game.
The most common type of BOs we use are smart entities:
BOs that embed intelligence into the environment.
We developed four kinds of smart entities: 
smart objects, navigation smart objects, smart areas and quest smart objects.
We also created situations as a very different type of BOs.
Figure~\ref{fig:objects_overview} shows the relationships between types of BOs we implemented.
and Table~\ref{tab:object_examples} gives examples of BOs that are implemented in the game.

This section introduces the 
AI system which formed a base for our code,
discusses smart entities and their variants and 
introduces situations.
An overview of our tool support and screenshots of the environment 
are given in the online supplemental material. % along with examples of BOs that are implemented in the game. 

  \begin{table}[!t]
  %   increase table row spacing, adjust to taste
   \renewcommand{\arraystretch}{1.3}
   % if using array.sty, it might be a good idea to tweak the value of
   % \extrarowheight as needed to properly center the text within the cells
   \caption{Examples of implemented behavior objects}
   \label{tab:object_examples}
   \centering
   \begin{tabular}{ll}
   \toprule
   \textbf{S-areas} & Pub, Shop, Field, City, \ldots\\
   \midrule
   \textbf{S-objects} & Chair, Bench, Bowl for eating, Door (navigation) \ldots \\
   \midrule
   \textbf{Situations} & Small talk, Beggar and a rich man, Collective dance, \ldots \\
   \midrule
   \textbf{Quests} & Bailiff's lost keys, Horse race, Bow training, \ldots \\
   \bottomrule
   \end{tabular}
   \end{table}

\subsection{Underlying AI System}

Our implementation is built on top of an existing AI system,
which is described in \cite{warhorseAI}.
The basic character decision making is performed 
by a variant of behavior trees \cite{champandard2007behaviortrees}.
The behavior tree (BT) formalism is extended with variables and 
a custom type system which allows for complex structured types and 
type inheritance.  

From the perspective of behavior injection, 
there are four very important aspects:
The first one is the extended execution model of the BT nodes
that ensures consistent initialization and cleanup of subtrees;
in particular it allows a specific cleanup behavior to be executed 
when the behavior is interrupted. 
This feature is critical to allow behavior objects and NPCs to maintain consistency (objective O\ref{problem:consistency}).
%The way a cleanup is expressed in the BT structure is shown in Figure~\ref{fig:behavior_cleanup}.

The second one is the hierarchy of \emph{subbrains}. 
Subbrains represent individual components of the NPC logic 
(ambient AI, combat, \ldots)
ordered by priority and connected in a manner similar to the subsumption architecture \cite{subsumption_brooks1991}.
If the subbrain becomes active, it executes a BT associated with it.
If a higher-priority subbrain tries to run, the BT of the lower-priority
subbrain is stopped, including proper behavior cleanup.
The subbrain priorities are fixed per NPC template and assigned by scripters. 
If more complex handling is required, the scripters may create a special ``switching'' BT running in parallel with
the subbrains and enforce activation/deactivation of specific subbrains through dedicated BT nodes.
The subbrain interaction is important for the situation BOs 
and to decouple combat logic and ambient AI in order to make sure that combat is always functional (objective O\ref{problem:gameplaycritical}).

% \begin{figure}
% \centering
% \includegraphics[width=0.7\columnwidth]{img/example_cleanup_tree}
%  
% \caption{Behavior cleanup example. The semantics of the RollBacker node ensure, that once the Clean Up branch is executed always}
% \label{fig:behavior_cleanup}
% \end{figure}

The third one is the possibility to create named links between game entities
and to query the link graph at runtime with complex queries 
(e.g., find all objects with ``child'' link to an object linked to me with a ``parent'' link).
We use this link system to connect BO instances with the environment data they need
(e.g., a pub is connected to all seats within; a quest is connected to an item the player should find).
This way instances have specific data, but the data is decoupled from behavior code (objective O\ref{problem:decoupling}).

The fourth one is a mature messaging system. 
NPCs and BO brains may have multiple \emph{inboxes}, each having its own
message type. The system manages thread-safe message queues for each inbox.  

% \subsection{General Remarks}
% We consider the distinction between on-request and on-command injection to be mostly conceptual. 
% At the implementation level (at least in our case) the distinction is not so clear cut, as the
% 
% Our first and in the end most complex implementations of behavior objects 
% are \emph{smart areas} --- behavior objects that encapsulate behaviors
% in a given location of the world (e.g., a pub). 
% We have further applied the same philosophy to strengthen the classical notion of \emph{smart object} 
% and also allowed those objects  be used as a part of NPC navigation.
% Smart objects and smart areas are similar in many aspects and we refer to them collectively as \emph{smart entities}.
% A similar encapsulation has also been applied to quest logic 
% and to \emph{situations} 
% --- short prescripted scenes involving multiple coordinated NPCs (e.g., small talk or a brawl).

\subsection{Smart Entities}

A \emph{smart entity} (SE) is a behavior object associated 
with a specific entity in the game 3D world.
We use two basic types of SEs --- \emph{smart areas} (s-areas), which
are associated with an area in the game world (e.g., a pub),
and \emph{smart objects} (s-objects), which are associated with a specific object (e.g., a door).
S-objects are further divided into regular, navigation and quest s-objects.

We will first discuss the common properties of SEs and later deal with their differences.
Note that while there are huge differences in the way different types of SEs are used in practice, %kandidat 
there are only minor differences in their implementation.
This provided significant speed-up of development and allowed reuse of the same behavior development tools.
 
SE contains all the information the NPCs need to behave 
appropriately in the context of a game entity 
(e.g., scripts for the innkeeper and for the guests, scripts for opening the door).
If necessary, the SE's brain 
coordinates various NPCs within 
(e.g., assigns free seats to the guests, chooses NPCs to engage in a brawl, handles queues near the door \ldots).
As such, SE together with the in-game entity form a standalone package 
that may be plugged in to the virtual world and be used by NPCs without modification to NPC code (objective O\ref{problem:decoupling}).
The SE may also disable certain behaviors and limit the maximal number of NPCs that may hold a given behavior at one time.

\subsubsection{Behavior Injection for Smart Entities}
To ensure that gameplay-critical behavior remains uninterrupted (objective O\ref{problem:gameplaycritical}), we have decided that
the injection should be performed on-request.
We have thus added a new BT node that requests a behavior from SE 
(further referred to as \emph{request node}). 

% Both the request node and the SE take part in deciding, which
% behavior is applicable.
% Let us consider the following example: an NPC enters a pub and asks for a ``have-fun'' behavior.
There are two possible configurations for the request node that are handled in a slightly different way:
1) the designer explicitly states the name of the behavior that should be requested
and 2) the behavior is left unspecified. 
In the former case it is checked, whether the SE provides a behavior with the given name, whether the behavior is currently enabled 
and that the maximal number of holders for this behavior is not exceeded.
If any of the conditions is not satisfied, the SE returns failure and the request node fails.
% It may further check constraints on requesting NPC   
% (e.g., a template that involves inviting  everybody for drink is available only to NPCs that are marked as rich).
% If the conditions are not satisfied, the pub returns failure and the request node fails in turn.
% For a more complex usage, it is also possible that the NPC specifies a list of preferred behaviors
% and/or a list of forbidden behaviors and an applicable behavior is chosen to honor these constraints
% (e.g., it may prefer ``get-drunk'' template to ``eat'' or forbid ``play-cards'' template).
In the latter case, the first available behavior is chosen,
failing only if there are no behaviors available.
This is used especially in the context of s-objects which often provide only one behavior. 
Another use is for the special case, when the NPC wants a smart area to just give it any behavior, 
which is useful for ``idle'' behaviors. This is specific to ambient AI (objective O\ref{problem:ambient}).

% There used to be more complex logic available for requesting behaviors 
% (the designer could give lists of preferred and forbidden behaviors).
% However it made SE debugging challenging --- it was often unclear, 
% why the request node failed to obtain a behavior.
If a more complex logic for requesting behaviors is necessary, the NPC requests a high-level behavior from the SE.
The high-level behavior then conditionally requests more specific behaviors from the SE.
This approach is used primarily when NPCs should behave differently in the same context based on their traits,
preferences or knowledge (objective O\ref{problem:different}).
% While this makes the BTs grow a little in depth, it enables the scripters to use the same debugging
% tools they use for BT debugging which speeds up development.
% If the tree depth became a performance issue, it would be possible to inject the specific behavior
% at the original request node instead of nesting it deeper in the tree.

If the requested behavior is available, it is instantiated in a data structure called \emph{behavior descriptor} 
which is passed to the request node.
The behavior descriptor contains meta data about the behavior (e.g., when it should be dropped) 
and an instance of a BT that achieves the behavior, which we call \emph{injected subtree}. 
The injected subtree is then added as the only child of the request node and the tree continues execution
by evaluating the subtree.
The injected subtree has access to the NPC's state and data and thus may modify the behavior appropriately 
(e.g., a rich guest in a pub orders more expensive food --- objective O\ref{problem:different}).

If needed, the behavior descriptor contains new message inboxes that should be added to the NPC
to allow synchronization and communication (objective O\ref{problem:synchronization}).
%The request node is responsible for subsequently removing those inboxes when the behavior is dropped. 
% The AI system supports nested inbox contexts in order to avoid problems with name clashes, 
% i.e., if the behavior descriptor contains an inbox which has the same name as one of the inboxes the NPC already possesses,
% references from behaviors defined in the SE will resolve to the newly added inbox 
% while references from outside will resolve to the original inbox.
% This is necessary to maitain consistency (objective O\ref{problem:consistency}). 

As synchronized action of multiple holders is often required (objective O\ref{problem:synchronization}), 
the descriptor also refers to the SE's local context in which locks are resolved (the context is part of the instance state).
This ensures that using a fixed lock name across multiple SE instances is safe. 
For example, when NPCs sitting around a table (a BO instance) in the pub want to synchronize movement during a toast,
they may all explicitly reference ``toast'' lock. 
Since the lock name is resolved relative to the BO instance, holders of the same behavior at another table instance will 
receive a different lock when referencing a ``toast'' lock. 
This improves code readability and prevents the necessity to share a lock explicitly by messages. 

As requests may be nested, it is technically possible to request the same behavior twice from the same object.
But this is considered a runtime error, as the expressive power of recursion 
would do more harm than good in a game setting.

% \begin{figure}
%   
%   \begin{algorithmic}
%    
%    \Procedure {IsInjectable} {$template$,$npc$}
%    		\If { $template.enabled$}
%    			\State \Return false
%    		\EndIf
%    		\If {$template.numInstances >= template.max$}
%    			\State \Return false
%    		\EndIf
% 		\If {\Call{ConditionMatch}{$template.condition$, $npc$}}
% 			\State \Return true
% 		\Else
% 			\State \Return false
% 		\EndIf
% 			   		
%    \EndProcedure
% 
%    \Procedure {RequestSpecific} {$behaviorName$,$npc$}
% 		
% 	\EndProcedure
%   
%   \end{algorithmic}
% 
%   \caption{The algorithms for behavior injection.}
%   \label{alg:injection}
%  
% \end{figure}
   
\subsubsection{Smart Entity's Brain}
   
The basic decision making of the SE is passive:
for each behavior, the SE maintains information
whether it is enabled (i.e., whether new instances of the behavior may be requested)
and the maximum number of instances that may be adopted at the same time. 
This information is used upon request processing.

Some SEs, especially areas, however need to have brains to actively influence the behaviors. 
The brain contains a behavior tree that gets updated regularly and may
either modify the passive decision making based on external conditions
(e.g., disallow ``drinking'' behavior in a pub if no innkeeper is present)
or it may perform some coordination among behavior holders inside the area
(e.g., instruct a pair of customers to play cards together).
The coordination is done by sending messages between the SE and the behavior holders.
Since the NPCs are now controlled by the injected subtrees, the SE can make
strong assumptions about NPCs responses to its messages.
Even if the NPC terminates the injected subtree, 
the cleanup logic of the behavior will notify the SE of this fact and allow for recovery.
This central control of joint actions is an important aspect of our implementation
as it removes the need for NPC negotiation (related to objective O\ref{problem:synchronization}).

There are special BT nodes specific to the SE brain that enable/disable behaviors and that send messages to %kandidat
holders of a certain behavior. 
The brain BT can access variables containing references to behavior holders 
and system data (e.g., what behaviors are enabled).
The BTs for the NPC behaviors then may use a special node to send messages to the SE that the NPC received behavior from. 

In many scenarios, the SE needs to perform some action 
whenever an NPC adopts a certain behavior (e.g., assign a free seat to a customer in a pub)
or when an NPC drops the behavior (e.g., innkeeper says goodbye to the leaving guest).
To streamline the development in such scenarios and to make the BTs of the SE brain and
the behaviors more readable, we have introduced event handlers to the SE brain.
An event handler is simply a BT that is executed until completion for each instance of an event.
%This further reduces the complexity of the code.
  
All of the SEs implement two events OnAdopt --- an NPC adopts a behavior --- and OnDrop --- an NPC drops a behavior.
S-areas introduce two more events that fire whether the NPC has requested a behavior or not: 
OnEnter --- an NPC enters the area --- and OnExit --- an NPC leaves the area. 
 
The SE adds events to an event queue. 
If the event queue is non-empty upon updating the SE, 
the handler tree of the event to be processed is updated instead of the main tree.
In order to keep handler code simpler and without safety checks and to simplify debugging,
the handler trees are executed without interruption.
The designers however must make sure that the handler trees complete quickly.
Our current practice is to only update the state of the SE or send messages inside the handler trees 
and perform any actual actions on the main tree. %, using the state/messages.
To prevent the main tree from starving  
at least one update to the main tree is guaranteed between two successive events.
% The algorithm for SE update is given in Figure \ref{alg:se_update}.  
% 
% \begin{figure}
%   
%   \begin{algorithmic}
%    
%    \Procedure {UpdateSEBrain}{}
%    		\If { event queue is not empty}
%    			\State $currEvent \gets$ head of the event queue
%    			\If {$currEvent.tree$ is finished}
%    				\State remove $currEvent$ from the event queue
%    				\State \Call{UpdateBT}{$mainTree$}
%    			\Else
%    				\State \Call{UpdateBT}{$currEvent.tree$}   				
%    			\EndIf
%    		\Else
% 			\State \Call{UpdateBT}{$mainTree$}
%    		\EndIf
%    \EndProcedure
%    
%   \end{algorithmic}
%    
%   \caption{The algorithm for a single update of a SE brain.}
%   \label{alg:se_update}
%  
% \end{figure}

\subsubsection{Linking Data to Smart Entities}
As there is an in-game entity (e.g., pub area, chair 3D model) attached to every SE instance, it is possible to use the 
linking feature of the underlying AI system to connect the instance to its environment data.
This is easily done and visualized in the game editor.
For example, the pub area has a link labeled ``seat'' to all chairs available for guests in the particular pub
and further labeled links for the beer tap and other notable locations in the area.
Upon initialization, the SE gathers the immutable environment data from the links to its internal variables to simplify
access.

\subsection{SE: Smart Objects} 

S-objects are SEs with the simplest intended use. 
They mostly handle short behaviors associated 
with specific in-game objects 
(sitting on a chair, opening door, cooking on a fire, \ldots).
The environment data of s-objects is usually only the 3D model they are attached to.
To reduce system load, majority of s-objects do not have their own brain and act only passively.

Still most of the s-object behaviors cannot be reduced to animations only,
because we aim for high behavioral fidelity.
For example, when sitting on a chair, we want the NPC to  
move the chair a little away from the table with its hand, 
go closer to the table and drag the chair back to its original position while sitting down.
To properly align the chair with the NPC, it must be first attached to the NPCs hand by its back, 
then the hand is detached and later the NPCs bottom is attached to the chairs seat.
Without the attachment, the chair might easily become slightly out of sync with the NPC
producing an eerily looking result --- this is a limitation of game engines in general 
and cannot be easily overcome.
Since the s-object provides complete code and not only an animation, we are able to handle these issues easily.
% The chair should also be able to provide multiple starting points for the NPC so that 
% the NPC does not go to the opposite side of the chair to sit.

Although our s-objects are used for the simplest use-cases in our game,
they are still much more powerful than s-objects 
in other OWGs that we know of. 
Apart from the simple uses outlined above,
more complex scenarios are supported.
The most elaborate s-object we have deployed so far is a bench that allows up to 4 NPCs to sit on it.
Since the bench is attached to a table, NPCs cannot stand up directly, but need to move to the 
end of the bench and then leave.
If an NPC in the middle wants to go away, the NPC on the side stands up, clears the way
and then sits back again.   
 
\subsection{SE: Navigation Smart Objects}

Navigation s-objects are an extension of regular s-objects. 
Navigation s-objects work as a link
between two navigable areas that would be disconnected otherwise. 
The most common ones are doors or barriers that can be jumped over.
The purpose of the navigation s-object is to provide a behavior that the NPC should use to
traverse the link.
As in the regular s-object case, we needed a more powerful mechanism than just playing animations.
A good example is a door: not only does the NPC play an animation, it also must be properly synchronized
to the door and, more importantly, a queue of NPCs waiting for the door must be handled reasonably.
For this purpose, the doors are linked to nearby places where NPCs should wait for their turn in the door
and explicitly manage the queue, including giving way to the player. 
We have chosen this central approach in favor of distributed solution using steerings or similar techniques
because given the specifics of our AI and animation systems and various minor design requirements,
the central control allows for much better results, although with some extra work.
%Once again, all this must work well when interrupted. % which makes things a little more complicated.

Navigation s-objects also differ from regular s-objects in injection method.
The navigation s-object behavior is injected on-command ---
the NPC does not partake in the decision to use the s-object, it is the navigation system that decides that the
particular s-object is used during movement. 
The injected subtree is then inserted as a child of the move node and updated accordingly.
Once the injected subtree finishes, the move node resumes its normal execution if the subtree was successful
or fail if the subtree failed.
The injected subtree is removed from the move node in both cases.

\subsection{SE: Smart Areas}

Smart areas (s-areas) are smart entities connected to whole areas in the game world, 
which has several implications for the way they work.
Most notably, s-areas capture higher-level behaviors than s-objects.
Often the s-area delegates the low-level functionality to s-objects. 
In the pub example,
the chairs or benches in the pub area are s-objects that provide ``sit'' 
behavior that is then requested from the ``guest'' behavior.
A typical s-area thus has a large amount of environment data 
which consist mostly of smart objects it uses.

\subsubsection{Behavior Requests and Smart Area Hierarchy}
%There is a difference in requesting the behavior in s-areas and s-objects:
In contrast to s-objects, where the NPC has to possess an explicit reference to an s-object instance while requesting a behavior,
an s-area may be used implicitly as ``the s-area I am currently in''.
While in some cases it turned out to be more useful to use s-areas with explicit references as well,
implicit area referencing is useful when the NPC wants to perform any particular behavior from a larger group of behaviors.
 
A typical example is ``relax'' behavior: the NPC wants to perform any relaxing activity
that is available in the area it is in (e.g., drinking or dancing in a pub, idly resting at a field or watching comedians in the city center).  
It is actually beneficial, if the relaxing activity is different when requested repeatedly.
We call behaviors where an NPC is not bound to a particular area \emph{general}. 
This is in contrast with \emph{specific} behaviors, such as ``work'' where the NPC has a specific place where it works
and this should be the same every time it works and thus an explicit reference should be used. 

There is however a catch in using general behaviors:
to reference an s-area implicitly, the NPC must be inside the area. 
But how does the NPC know, where a pub is if all it requires is to relax? 
As mentioned in our design objectives, the pub location 
should not be hardcoded in the NPC's behavior.
Our solution was to introduce parent-child relationship between s-areas
and make the whole city an s-area and make the pub its child. 
Now the city (the city designer) knows the locations of all pubs and other relaxing areas within.
The NPC thus requests a ``relax'' behavior implicitly 
and the city s-area gives it a BT that consists of a sequence of a move node that moves the NPC
to the pub and a request node that requests a ``drinking'' behavior in the pub.

\begin{figure}
\centering
\includegraphics[width=0.85\columnwidth]{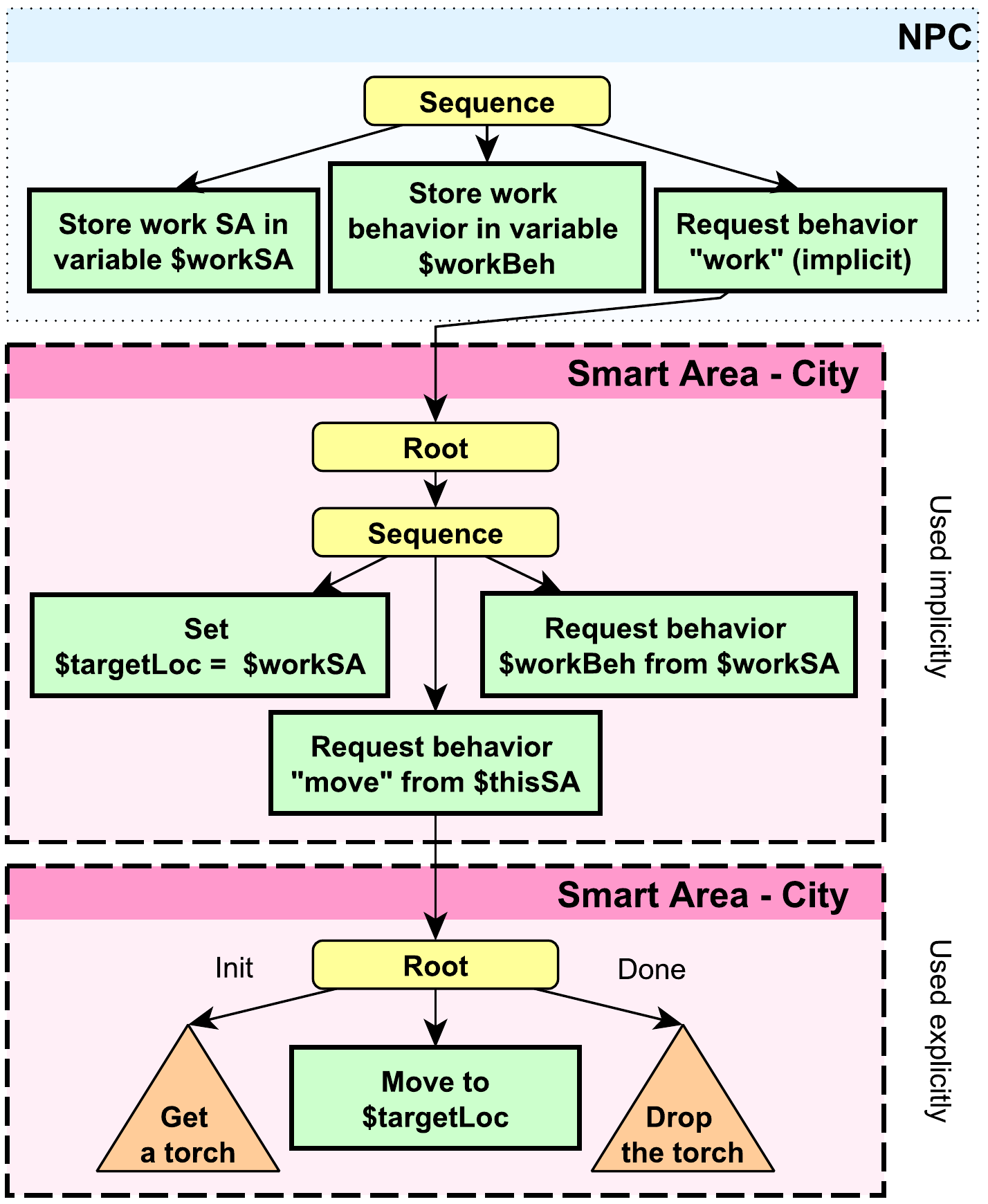}

\caption{Handling specific behaviors at high-level smart area to control movement within the area.
The \$thisSA variable is provided by the system and refers to the s-area that provided the behavior.
Note that when s-area behaviors are nested, \$thisSA will refer to a different value in different subtrees.}
\label{fig:move_behavior}
\end{figure}

% Initially, we tried all s-area behaviors to be general,
% but this turned out problematic, as the city had to be able to direct NPCs to particular places where they work or where they sleep,
% although all the information for this decision had to be already connected to the NPC 
% introducing unnecessary coupling. 
However it later proved necessary to involve higher-level areas in specific behavior execution as well.
The reason behind this is that the s-area should be able to control or modify all movement of NPCs within its bounds 
(e.g., make the NPC pick-up a torch when moving at night).
A high-level s-area is a good place to store this kind of behavior as it applies globally to all movements within the area.
The code for this move behavior is generic and not bound to any particular behavior 
(the target location is passed to the injected subtree through a shared variable, 
because our system currently does not support parameters to behavior requests).
An example of injection of this kind of movement behavior is shown in Figure \ref{fig:move_behavior}.

As of now, sleeping, eating and most of working behaviors are specific, 
but the pastime behaviors (fun, prayer, \ldots) 
and some non-distinctive working behaviors (e.g., fishing or hunting) are general.

A different problem we aimed to solve with s-area hierarchy 
is that an NPC that is currently in a pub may decide it wants to pray, but the pub
should not be required to know of all churches in the city.
It is thus a good idea to ask the city in such a case.
For this reason,
if the current s-area cannot provide any applicable behavior, the request node asks the parent s-area.
To avoid confusion, we have adopted the practice that behaviors in leaf areas
have distinct names from behaviors in the parent area 
and for general behaviors, only the behaviors from the higher-level areas are requested.
This way, the higher-level areas always take part in decisions about general behaviors
that take place within its bounds and may for example balance the amount of NPCs in
individual pubs in the city.

There is however one possible exception, when defining the same behavior in both child and parent areas %kandidat
might be desirable. 
This would be the case with general behaviors like ``go to toilet'', where the NPC should stay in the same s-area,
if it may perform the behavior there, but should be able to ask a parent area, if this is not possible.
As of now, we have however not identified any such behavior that the design team would want to have implemented in the game.

We have also introduced a default top-level area, covering the entire game world.
This way general behaviors can be requested anywhere on the game map and
the default area is able to guide the NPC to an s-area that provides such behavior.

% To provide NPCs with behaviors in reaction to dynamic events the system is prepared to support
% s-areas created on the fly. 
% For example, after a murder on the streets an s-area that instructs the witnesses to run away should be created.
% Although this is conceptually an on-command injection, 
% we plan to implement it using the same technique as with regular s-areas to keep the system implementation simple: 
% our current working concept is that such an s-area will send a message to all NPCs within its bounds. 
% This message will activate a higher priority branch in the NPC's behavior tree or a specific high priority subbrain 
% which will contain a request node asking for relevant behavior.

\begin{figure}
\centering
\includegraphics[width=0.8\columnwidth]{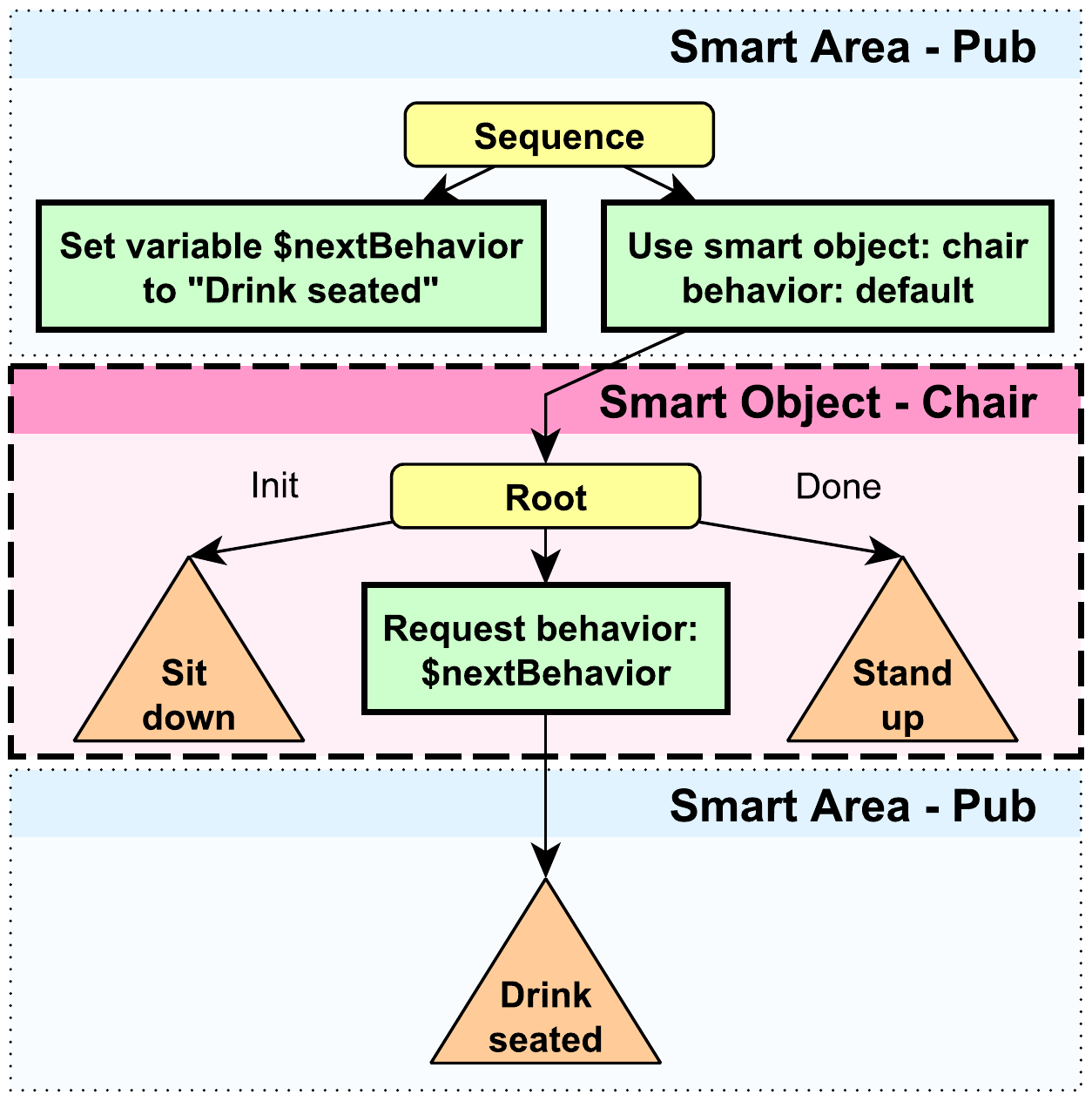}
 
\caption{Using smart object to ``decorate'' a behavior in a smart area. 
In actual implementation, the node structure at the s-object behavior root is a bit more complicated, 
but is conceptually equivalent to the structure in the figure.}
\label{fig:sa_smart_object}
\end{figure}

\subsubsection{Using Smart Objects Inside Smart Areas}

One of the interesting problems we tried to solve was how to properly use
s-objects, in particular chairs, inside s-areas.
The typical problem is as follows: the pub s-area wants the NPC to sit down, 
wait for a beer and drink it, then stand up.
While sitting down and standing up should be delegated to a s-object,
it is necessary that the behavior in between remains controlled by the s-area
while still letting the s-object make sure, that the NPC does not remain seated 
if the behavior is terminated.

To keep the s-object in control of init/done behaviors, 
the solution we adopted is that the s-area behavior requests the
s-object behavior which in turn requests a ``private'' s-area behavior that expects the
NPC to be seated.
The name of the private behavior is passed through a shared variable.
%Technically, we want the chair behavior to work as a decorator node over the private s-area behavior.
An example of the setup is given in Figure \ref{fig:sa_smart_object}.

\subsection{SE: Quest Smart Objects}
\label{sec:qeust-so}

We divide quests into two categories with regard to the way they affect NPCs:
quests that do not require NPCs to change their behavior are called
\emph{behavior-preserving} and those that do are \emph{behavior-changing}.  
An example of a behavior-preserving quest is a guard asking the player to kill bandits living in the woods. 
The bandits behave the same way they would without the quest (attack the player when they see them)
and the guard keeps guarding the village. 
The only change is in dialog options the guard provides 
--- once the bandits have been killed a new dialogue to congratulate the player on success in the quest is activated.

The quest system in our game is event-based and handles behavior-preserving quests very easily.
A quest is composed of a series of steps.
Individual steps of an active quest listen to events in the game (player picking up items, killing enemies, \ldots).
These events then trigger progression of the quest to next steps.
The quest steps are also bound to dialogues that the NPCs use and allow/disallow
various dialog options.

Behavior-changing quests are however more demanding.
We did not want NPCs participating in a quest to completely abandon their daily cycles and
stand at one place, waiting for the player. 
Instead, behavior changing quests may directly modify daycycles of the participating NPCs.
Initially we envisioned a new type of behavior object to encapsulate behaviors related to a given quest,
but to save development time, we decided to use \emph{quest smart objects} to handle this task.
A behavior-changing quest delegates execution of some of its steps to a quest s-object. 
The quest s-object then notifies the quest of completion/failure of the assignment by the player.
This communication is done through the message system.

The quest smart objects are technically the same as regular s-objects 
which allowed us to directly reuse code for both the game and the editor,
but they are used very differently.
The most notable difference between regular s-objects and quest s-objects is that quest s-objects are connected to \emph{quest anchors} 
--- game entities not visible in the game.
Quest anchor's only function is to connect the quest s-object to its environment data.
Quest s-objects also always have brains that guide the execution of the quest step(s).
The quest s-object then may instruct the NPC to exchange a part of its daycycle with a behavior requested from the quest s-object.

If the quest requires the NPC to change its behavior completely, regardless of the day cycle, %kandidat
the appropriate approach would be to introduce new higher-priority quest subbrain to the NPC.
The quest s-object would then activate the quest subbrain which will in turn request a behavior from the quest s-object.
However all of the quests implemented so far are designed to keep at least the basic daycycle intact (in particular let the NPC sleep at nights).
%this was not yet necessary and we have not implemented it so far.

\subsection{Situations}

A situation is a very different kind of BO than smart entities.
It encapsulates a short coordinated behavior involving multiple NPCs. 
Typical examples of situations in our context are two villagers pausing for a small talk,
a collective dance in the pub or a brawl.
An important aspect of situations in our implementation is that they serve
mostly as ``eye candy'', i.e., they should not significantly alter the state of the game world.
This is important because it lets the AI system run situations without considering
their consequences for the current state of the game.
Situations have deliberately very low priority so that any ``important'' behavior
always overrides the situation. 

Technically, situations are run within a specific situation subbrain of the NPC, 
which has a higher priority than ambient AI, 
but lower priority than any other subbrain.
Thus if only an ambient behavior is being performed, 
and the NPC should start performing a situation, the ambient AI is suspended 
and the behavior given by the situation is started.
This could still lead to undesirable results (such as an NPC starting to dance in the middle of a conversation).
To keep the ambient AI in some control over situation execution,
the NPC has to explicitly subscribe to the situation system. 
This is achieved by decorating a subtree of the ambient AI behavior with a special node that
subscribes the NPC when the execution of the subtree starts and unsubscribes when the execution is finished.  
This makes it possible for situations to be developed almost independently of the rest of the AI code,
as the potential negative interactions with other behaviors are minimized by design (protecting gameplay-critical behaviors - objective O\ref{problem:gameplaycritical}). 

A situation template describes several roles, each providing a behavior for one of the NPCs that
participate in the situation.
Roles also have associated conditions that an NPC must satisfy to take the given role
(e.g., to engage as an aggressor in a pub brawl, the NPC must be drunk).
The situation templates are connected to s-area templates to allow for area-specific situations.
Once a component called situation manager decides that a particular 
situation template should be instantiated, 
it tries to find suitable NPCs using constraint satisfaction techniques.
After the NPCs are chosen, an instance of the situation is created and the behaviors are injected on-command as the 
main tree for the situation subbrain making the subbrain active.
%\footnote{
%To avoid multithreading issues and inconsistencies, the behavior is sent to the situation subbrain as a request for injection.
%The situation subbrain then injects the behavior during its next update.}.
More details on the inner workings of the situation manager and the whole situation subsystem 
are out of scope of this paper and can be found in \cite{situationsystem}. 

If any of the chosen NPCs cannot execute the situation or terminates the situation prematurely 
(e.g., because a higher-priority subbrain becomes active), 
all other participants also abort the situation.
Once again, the clean-up behaviors are guaranteed to be executed, keeping the system in a consistent state (objective O\ref{problem:consistency}).
After all participants finish their behaviors, the situation instance is destroyed.

As a behavior object, situation is lightweight compared to SEs.
This is mostly because situations are much more specific than SE behaviors 
and that situation instances are short-lived.
In particular, situations do not have their own brain,
as central decision making is usually not necessary
and in the rare cases when it is, one of the holders may handle the central logic.

For coordination purposes (objective O\ref{problem:synchronization}), all participants are given explicit references to all other 
participants and a local synchronization context is maintained for the situation.
The situation also provides the participants with up-to-date information
on the state of the other participants, especially if they already started the given behavior
or if they dropped the behavior and thus may no longer be expected to cooperate. % react to other participant's requests. 

\section{Evaluation}
\label{sec:evaluation}

The evaluation in this paper consists of two main parts: 
qualitative observations gathered in 16 months since 
the scripters first used smart areas for development 
and two rounds of semi-structured interviews we performed with scripters.
% Although not backed by empirically large samples  
% we think that the lessons learned from applying the technique in production are an important
% contribution of our paper, 
% since evaluation of academic techniques in real-world game development is very seldom performed.
In our previous work, we have also performed quantitative evaluation, 
which we sum up before the new results in this paper.   

\subsection{Summary of Evaluation in Previous Work}

In our previous work, we have performed quantitative evaluation 
of scripters' performance using BTs and using BTs with s-areas \cite{cerny2014smartareas}.
While part of the measured metrics showed statistically significant differences,
the sample size and the scope of the tasks assigned to the participants was limited
and we concluded that the ``data provide some support that s-areas are better,
when modifications are frequent --- which is the case in real development ---
but the results are not clear and further research is needed.''
We have however not managed to perform a new quantitative study for this paper.  

Qualitative feedback was also gathered, including the fact that
``subjects were relatively quick at understanding code created with s-areas.
Judged by the researcher monitoring experiment progress, 
the subjects using s-areas had no trouble finding the code for a particular behavior [in contrast to subjects not using s-areas].''

We have also tested the computational load the AI system --- including SEs ---
imposes on the CPU and found that it is fast enough for production \cite{warhorseAI}.
The system consumes less than 1~ms on average with 30 complex NPCs running and less than 2 ms with 300 simple NPCs running.
However, the large amount of s-objects turned out to be a bottleneck.
In response we decided to reduce the number of s-objects that have their own brains
and to update those that have brain less frequently to ensure swift execution even with larger worlds. 
BO instances also share pools of instantiated BTs and inboxes to reduce memory footprint 
while keeping the performance benefit of preallocated and preconstructed objects. 

\subsection{General Observations}

In general, the virtual world works well using BOs and the structuring of behaviors
into objects lets scripters concentrate on individual aspects of the world (pub, shop, church, \ldots)
while only minor problems arise during integration of the individual objects into the world (objective O\ref{problem:decoupling}).
As an example, one of the scripters was recently tasked with adding ambient AI to a freshly created village.
This required placing approximately 60 s-areas and 200 s-objects in the world, linking them to environment data 
and performing basic tests. The scripter completed all those tasks by himself in two days.

% We have been able to implement behaviors that --- to our knowledge ---
% have not been present in any commercial game, for example realistic door and seating logic, complex interactions with items in a pub, \ldots) 
% and those behaviors were included in the public release of an alpha version of the game. 
S-areas and all types of s-objects are considered stable and have been deployed in a public alpha version of the game. 
The situation system is still in preliminary use but unlikely to change significantly.

We have also noted that s-areas, quest s-objects and situations closely
correspond to the way game designers think about the world:
it is natural for them to describe the behaviors that NPCs should manifest in a pub
separately of other behaviors the NPC perform in their daily cycles.

So far, 30 types of s-areas and over 40 types of s-objects 
of release quality have been developed and released in the public alpha version.
Nine situations were developed to test the situation system, 
but these situations will be subject to heavy changes before inclusion in a release build. 
7 quests have been released in the public alpha version and multiple others have been developed in release quality.

% Several difficulties have also been observed.
% One interesting problem arised during experiments with s-objects that may be picked up by the player or the NPCs.
% The s-object then contains code the NPCs as well as the player avatar execute to perform actions on the object
% (e.g., food providing specific eating animation, long fishing rod providing specialized pickup/put down animations)
% and possibly some data, for example the place the NPC should return the object to after use.
% When multiple instances of the same item are present in the inventory,
% because the inventory system expects two instances of the same item to be completely interchangeable,
% which is no longer the case as there are some instance-specific data.

We received mixed feedback to the fact that s-areas have strict boundaries. 
Boundaries introduce issues to handle when the movement to the area fails for some reason
or produces unnecessary movement, in case the s-area instructs the NPC to leave the area (e.g., to gather wood outside the area).
On the other hand, strict boundaries are beneficial from debugging perspective --- one can be sure, 
that since the NPC is outside the area, it cannot receive a behavior from it.

% As the system is new, no coding standards or best practices were known beforehand and could not be enforced.
% Through many rounds of prototyping of both s-areas and s-objects,
% the code of individual scripters has noticeably diverged.
% For example, the table where NPCs eat at home has different code than the table in the pub,
% because it was made in different context by different scripters and they provide conflicting interface 
% to the parent s-area.
% This is however to be expected, as the same issues arise in any form of programming, 
% and the remedy is also similar --- enforcement of common best-practices and reusability.
% However, having diverse solutions for the same problem was actually useful several times 
% as it allowed to test multiple variants and choose the best to form the standards 
% that can be enforced later on.  
 
Another lesson learnt is that it is vital to keep the s-object behaviors small and focused on a single task
while providing detailed control to the parent s-area.
An example that used to be problematic is feeding fire in a house.
In an initial implementation, the house s-area told the NPC to use a fire s-object.
If the fire s-object realized there is no wood, it instructed the NPC to use an s-object representing a pile of wood outside the s-area.
Now the s-area believed that someone was performing fire feeding behavior and should be done quickly, but in fact, 
the NPC was outside the area on a much lengthier task.
Our current implementation is that if there is no wood, the fire feeding behavior fails. 
The s-area is notified of the reason for the failure and assigns a ``find wood'' behavior to the NPC. 
This way the s-area is more aware of what is going on and may react to individual events.

The above example also illustrates where the system evolved to:
the individual behaviors are kept small and are hierarchically requested
from a large number of BOs.
One further example is the pub. 
The pub directs NPCs to table s-objects which in turn delegate the actual sitting to several attached chair s-objects.
The table also manages a bowl s-object that manages pieces of chicken (also s-objects).
To eat, the NPCs thus request eating behavior from the table, which requests behavior from the bowl,
which requests behavior from the chicken pieces.
This arrangement is instructive and scripters are happy that building a new pub can be done
by simply arranging the premade s-objects and connecting them with links.
Further, any element can be replaced without changes to the others (e.g., a plate instead of a bowl, a piece of pork instead of chicken) 
and the individual behaviors are easy to debug.
The downside is that the abundance of s-objects is taxing on the system by both the need to manage
the s-objects and by making the NPCs trees deeper and thus slower to evaluate.
However, this load seems manageable so far.
    
% Another performance related technicality it that it was very important to
% pool and reuse the injected subtrees instances instead of creating and destroying them on the fly.

\subsection{Qualitative Feedback}

We performed two rounds of semi-structured interviews with all 6 scripters
currently employed by the company. 
Except for the technical design lead, 
these are all of the company employees that use BOs on a daily basis.
 
We have chosen a qualitative approach because
there are few scripters in the company and thus quantitive conclusions would be weak. 
The experiences of the individual scripters are also not comparable, 
as the scripters specialize and solve very different problems.
Recruiting external subjects to extend the sample size
is not practical, as a large amount of knowledge has to be mastered, 
before a user is able to deal with tasks at least remotely connected to actual practice.
Structured interviews have their limitations,
nevertheless they have given us valuable insights for further development of our variant of BOs
and we consider them of interest to anyone trying to implement their own.

Shneiderman and Plaisant \cite{userInterfaceBook} recognize five basic usability measures:
time to learn, speed of performance, rate of errors by users, retention over time
and subjective satisfaction.
We focused on subjective user satisfaction as this is the only category where we can,
to some extent, separate the effects of using BOs from the features and quirks
of the underlying AI system and BT implementation.

The first round of the interviews consisted of broadly formulated questions 
on the general usage of the AI system (including BOs), 
while the second round had more focused questions linked to the design objectives of BOs.%
% \footnote{As noted by one of the reviewers, the actual questions could have been
% more focused if we followed some of the established guidelines from human-computer interaction research. 
% We were however not able to redo the evaluation with a better experiment design and present the results ``as-is''.}

\begin{table}[!t]
%   increase table row spacing, adjust to taste
 \renewcommand{\arraystretch}{1.3}
 % if using array.sty, it might be a good idea to tweak the value of
 % \extrarowheight as needed to properly center the text within the cells
 \caption{Questions in the first round of interviews. The reason why we included the individual questions are shown in italics.}
 \label{tab:questions_1}
 \centering
 \setlength{\tabcolsep}{2pt} 
 \begin{tabular}{lp{0.9\columnwidth}}
 \toprule
Q1 &  What were the tasks you worked on recently? 
\textit{Frame the interview and provide source for specific examples for the rest of the interview.}\\
Q2 &  What activity consumes the most of your development time?
\textit{Discover the main bottlenecks for production.} 
\\

Q3 & Give an example of a code segment/snippet that is often repeated across behaviors 
and has to be copied each time and a segment that is well reused across behaviors. 
\textit{Discover a situation where BOs are not applicable in practice, although they should be in theory.
Understand the potential for AI code reuse.}
\\
Q4 & Describe the process of implementing a behavior from a design request to the final code.
\textit{Discover how BOs fit (or do not fit) in the overall production pipeline.} %kandidat na zkraceni
\\
Q5 & How would your behavior code change if you could only use plain tree injection (without BOs).
\textit{Understand what features of BOs are considered important.}
\\
Q6 & What was the most complex/difficult task you have worked on in this company?
\textit{The most challenging tasks are likely to demonstrate the full power (or lack thereof) of a system.} 
\\

Q7 & Describe the process of resolving an issue reported by the QA department.
\textit{Discover whether BOs help/hinder debugging.} 
\\

Q8 & What do you dislike about the scripting tools?
\textit{Gather all the problems scripters face when writing code.} 
\\

Q9 & Describe your ideal scripting tool.
\textit{Gather constructive suggestions and let the scripters compare BOs to hypothetical alternatives.} 
\\

 \bottomrule
 \end{tabular}
 \end{table}

\subsubsection{First round of interviews}

The interviews consisted of nine questions and took 30 - 60 minutes.
Table \ref{tab:questions_1} shows the questions and the information we expected to gather from the answers. 
In general, we tried not to mention BOs in the questions to make scripters more likely
to report when they used alternative solutions and to prevent bias.
%We also wanted to improve other parts of the AI system than BOs.
While we asked the scripters to report on the AI system as a whole,
we expected all the answers to reflect on BOs to an extent, as the vast majority of in-game behaviors 
are built with BOs. 

First, we will focus on issues with the system, which were mostly reported for questions 3, 8 and 9.
The scripters reported a number of usability problems with the underlying AI system,
especially with the BT editor and debugger,
but only three concerns that could be linked to BOs were raised. 
The most common issues related to BOs (mentioned by 4 scripters) were the
usability problems inherent in debugging large trees (e.g., ``the trees do not fit well on a single screen''). 
This is mainly an issue with our BT editor, but is related to BOs, because every injection adds depth to the tree.
%and our use cases require a large number of injections.
Second came the necessity to use global variables to parameterize injected behaviors (4 mentions by 2 scripters, see Figure \ref{fig:move_behavior} for an example).
One scripter also mentioned that he dislikes that
all s-objects need to be connected to an in-game entity, although for some quest s-objects
there is no natural connection.
Overall, BOs are seldom the source of frustration of scripters, although they are used on a daily basis. 
The answers also support our previous work \cite{gemrot2014methodology} where we show, 
that quality tooling support is vital for a technique to succeed in practice. 

Except for a few references to usability issues with the AI system, questions 4, 6 and 7 did not provide any
valuable insight into BO usage. 
The other questions however conveyed some interesting feedback.

The most time-consuming activities (Q2) were debugging in general and updating code 
after a backwards incompatible change has been made to the underlying AI system (both mentioned by 4 scripters).
Good debugging support is thus vital to a success of a tool. 
% It is also obviously beneficial, although not always possible in practice, to change
% the underlying AI system as little as possible during production.
One scripter reported development of synchronized behaviors as the most time consuming and one reported
that he spends most time in figuring out, how exactly should the relatively broad requirements from
game designers be implemented at the low level.

Examples given for good encapsulation (Q3) were very specific to our system and do not provide
a valuable insight into BO usage. 
However, 7 examples were given of frequently copy-pasted code.
All of those were small snippets consisting of up to 8 nodes and were not suitable candidates for BO-based
implementation  
(e.g., searching the link network for a useful object, aligning animations to game entities).
While this means that BOs let the scripters reuse larger code structures without problems, 
it also indicates a room for improvement of the AI system:
creating a reusable BT snippet should be made easy, especially it should be straightforward
to pass data (parameters) to an injected tree.  

Best insights into BOs were provided by Q5.
Scripters reported that without BOs they would reimplement:
the ability to connect behavior and data (4 mentions);
a local communication hub/an entity that handles messages related to a given context (3 mentions);
a central logic (brain) for a set of behaviors (2 mentions)
and
a container of related behaviors (1 mention).
One scripter also mentioned that BOs help him write consistent code 
and another stated that he ``would implement something very similar''.
We see that the defining properties of BOs (connecting code and data and
a centralized point for coordination) were
indeed perceived as important.

\subsubsection{Second round of interviews}
The interviews consisted of five questions which aimed to elicit feedback on how do BOs fulfill the design objectives of the system (see Table~\ref{tab:questions_2})
and took 10 - 30 minutes.

Q10 was bound to objectives O\ref{problem:gameplaycritical} and O\ref{problem:consistency}. 
Only one scripter reported that he has written complex code with interruptions in mind. 
He has been responsible for making behaviors work when interrupted with a dialogue. 
While it was not hard to let the NPC finish an uninterruptible task prior to dialogue,
main difficulties stemmed from the fact, that the behavior 
has to resume to the point where the dialogue started, while some of the NPC's state
resides only in the animation system, inaccessible to the BT. 
In particular, an animation may be queued for execution, 
but not actually started when the dialogue is invoked (see Section \ref{sec:shared_state} for the reasoning behind this).

To remedy this, an improvement in the animation handling was implemented,
letting the scripters to directly access animation state and to create ``lambda BTs'' 
--- a BT counterpart to lambda functions in classical programming languages.
Lambda BTs are subtrees that are attached to events in the animation system. 
These subtrees then get executed regardless of the progress in the main BTs and can send messages that are handled in an appropriate moment by the main BT.

 \begin{table}[!t]
%   increase table row spacing, adjust to taste
 \renewcommand{\arraystretch}{1.3}
 % if using array.sty, it might be a good idea to tweak the value of
 % \extrarowheight as needed to properly center the text within the cells
 \caption{Questions in second round of interviews. 
 The design objectives that motivated the questions are shown in italics.}
 \label{tab:questions_2}
 \centering
 \setlength{\tabcolsep}{2pt} 
 \begin{tabular}{lp{0.9\columnwidth}}
 \toprule
 Q10 & When writing code, do you take into account the possibility of interruption by quest/combat? How?
 \emph{(O\ref{problem:gameplaycritical} and O\ref{problem:consistency})}
   \\
 Q11 & Is there a difference in using BTs and BOs in quest logic and in ambient AI?
 \emph{(O\ref{problem:ambient})} 
 \\
 Q12 & What are the necessary steps to place a new instance of an s-area/s-object in the game world?
 \emph{(O\ref{problem:decoupling})} 
 \\
 Q13 & Have you implemented any behavior where the attributes of an NPC would change the way the NPC behaves in a given context?
 \emph{(O\ref{problem:different})} 
 \\
 Q14 & What was the most difficult synchronization/coordination task you implemented? Why?
 \emph{(O\ref{problem:synchronization})} 
 \\
 \bottomrule
 \end{tabular}
 \end{table}

The same scripter and three other colleagues have implemented simpler interruption-aware code that handled halting
of the subtree (stopping the behavior without the need to resume to the original state).
Two of those reported that it was easy and one other reported that 
BTs support halting well.

We see that the system demonstrates capability to properly handle ``hard'' interruptions
when the NPC discards the running behavior completely, 
while further refinements are necessary to the ``soft'' interruptions where the behavior is 
expected to maintain its state after the interruption has finished.

Q11 was intended mainly to check whether our focus on ambient AI (objective O\ref{problem:ambient}) has not introduced
problems in quest handling. 
This does not seem to be the case as no scripter reported 
notable  problems with writing quest behaviors. 
The only problem that was mentioned was the fact that quest logic intersects
with multiple systems with overlapping capabilities: quest s-objects, the dialog system 
and the quest system (see Section \ref{sec:qeust-so} for details).
The consequences are twofold: 1) writing quests requires the scripter to interact
with several different user interfaces and 2) there are multiple ways to distribute
the quest logic among the systems.
Our current consensus is that when a quest uses an s-object (i.e. when the quest alters behaviors of NPCs),
then all of the quest logic is implemented within the s-object and the other systems only pass messages to the s-object. 
   
Other than that, three scripters considered quest behaviors to be very similar to ambient AI 
and two scripters considered quest behaviors to be simpler in general than ambient AI.
One scripter has not implemented any quest yet. 
Although we did not ask directly about quest s-objects, two scripters said that quest 
behaviors differ in that quest s-objects serve as a central entity to coordinate the quest.
This indicates that quest s-objects do their job well.

For Q12, all scripters reported that to create a new instance of a BO,
they never needed to do more than link the BO to the appropriate environment data.
Two scripters explicitly said that the process was quick, while two reported 
on usability issues with the linking system. % that had slowed the work down.
This shows that behavior code is well decoupled from data and that 
objective O\ref{problem:decoupling} was fulfilled.   

With regards to different behaviors of NPCs based on their attributes (objective O\ref{problem:different}, Q13),
only one scripter has already implemented such a behavior.
This was a military camp, where soldiers are assigned different work tasks based on their rank.
He didn't report any problems in achieving this, but further investigation is still needed.
 
As for synchronization and coordination (objective O\ref{problem:synchronization}, Q14), all scripters encountered tasks that required
explicit synchronization of NPCs, but two only in a very simple context.
Only one scripter built synchronization outside the scope of a BO and 
he referred to this case as the most difficult to handle.
Another scripter explicitly mentioned that s-areas were helpful for coordination.
Three scripters considered synchronization to be non-problematic, 
while two reported usability issues 
with debugging and implementing synchronized behaviors.
One scripter also reported that he finds parallel behaviors challenging in principle. 
Two scripters reported usability issues with the message system that make writing 
message-oriented code tedious.
Two scripters described the need to reduce the scope of possible NPC states when coordinating behaviors
for quests --- when NPCs need to cooperate on a quest, they are usually
instructed to stay at a well-defined place and perform only very simple activities so that
other NPCs can make simplifying assumptions on their state.

One scripter reported a performance issue that arose while he was implementing advanced door handling behavior where NPCs 
form a queue, but there are dynamic priorities for NPC ordering (e.g., if the door is locked, an NPC that has a key is given priority).  
 This resulted in multiple rounds of messages being exchanged between NPCs.
As two-way communication cannot be performed within a single frame and there was
a relatively lot of computation involved between the messaging, the system exhibited visible lag
when many NPCs tried to use the same door at once. 
This will be resolved by both simplifying the code and by giving larger time budget to evaluate trees of s-objects that are heavily used.  

In general the data indicates that synchronization and coordination is handled by BOs in a satisfactory manner,
although improvements can be made, especially in tools and usability.

\section{Conclusion and Future Work}
\label{sec:discussion}
\noindent 

In this paper we have described behavior objects as a tool to manage script complexity
in OWGs. 
We have shown five use cases for BOs within the AI system for an upcoming AAA role-playing OWG.
The behavior object concept has withstood field testing and was deployed in multiple forms within a public alpha version of the game.
Qualitative feedback and lessons learned during the implementation were presented.
The available feedback suggests that BOs
are a suitable approach for managing complexity in NPC behaviors, 
fulfilling all design requirements.
Vast majority of the issues scripters experience when writing and debugging BO code
are usability problems and incomplete tooling support.
A simple improvement to be realized in short term is to provide more computational resources to heavily used BOs.

% These improvements will be implemented in the short term.  
% Further improvements are also necessary for correct resuming of interrupted behaviors with animations
% and to provide more computational resources to heavily used BOs.
% These improvements will be implemented in the short term.  
% We have described our implementations in great detail because our experience indicates that it
% is these details that are critical for successful industrial deployment.
% Thorough usability testing and right tool support have also been a vital
% part of the initial success of the technology. 
% We believe that these are lessons to be learned 
% by the broader academic AI public to promote adoption of techniques developed in academia by the industry. 

While our initial development of smart entities and situations 
was driven simply by the needs of the AI system, 
we have noticed the similarity of the concepts to object-oriented programming.
In this paper we have established the connection between behavior objects and OOP explicitly,
as it helped us drive further development of our particular implementation
and provided inspiration.
We believe that inspiration by OOP can be useful for the next generation of game AI
and lead to dramatic improvements in code manageability, as OOP has done for classical programming.
Our implementation is based on behavior trees, but BOs should be usable in
all reactive action-selection mechanisms that are in frequent industry use.

As a future work we plan to extend our BT formalism by letting injected trees have explicit parameters.
We also plan to use the BO concept in other AI components.
An example is a possible use of a variant of smart objects to drive combat events
(e.g., a table that allows the player/NPC to perform a special combo by throwing the opponent at the table).

% or


\section*{Acknowledgement}

This research is partially
supported by the Czech Science Foundation under the contract P103/10/1287
(GA\v{C}R), by student grant GA UK No. 559813/2013/A-INF/MFF and
by SVV project number 260 224.

Special thanks belong to Warhorse Studios and its director Martin Kl\'{i}ma for making this research possible
by their openness to novel approaches and by letting researchers work in close cooperation with the company.

% \ifCLASSOPTIONcaptionsoff
%   \newpage
% \fi



\bibliographystyle{IEEEtran}
\bibliography{behavior_injection}

%\newpage


%\vfill


\end{document}